\documentclass[letterpaper, 10 pt, conference]{ieeeconf}  

\IEEEoverridecommandlockouts                              

\usepackage{hyperref}       
\usepackage{url}            
\usepackage{booktabs}       
\usepackage{amsfonts}       
\usepackage{nicefrac}       
\usepackage{microtype}      
\usepackage{xcolor}         
\usepackage{bm}
\usepackage{algorithm}
\usepackage[noend]{algpseudocode}
\usepackage{graphicx}
\usepackage{subcaption}
\usepackage{wrapfig}
\graphicspath{ {./figs/} }

\usepackage{amsmath, amssymb}
\usepackage{enumerate}
\usepackage{tikz}
\usepackage{multirow}

\usetikzlibrary{shapes.geometric}
\newcommand{\myignore}[1]{}

\newcommand{\inbrace}[1]{\left \{ #1 \right \}}

\newcommand{\set}[1]{\inbrace{#1}}

\def \F {\mathbf{F}}
\def \G {\mathbf{G}}

\def \U {\mathbf{U}}

\def \X {\mathbf{X}}

\newcommand{\tuple}[1]{\langle #1 \rangle}

\DeclareMathOperator*{\argmax}{arg\,max}

\makeatletter
\newcommand*{\rom}[1]{\expandafter\@slowromancap\romannumeral #1@}
\makeatother

\title{\LARGE \bf
LTL-Transfer: Skill Transfer for Temporal Task Specification
}

\author{Jason Xinyu Liu$^{*1}$, Ankit Shah$^{*1}$, Eric Rosen$^{1}$, Mingxi Jia$^{1}$, George Konidaris$^{1}$ and Stefanie Tellex$^{1}$
\thanks{$^*$Equal contribution. $^{1}$ Brown University.}
}

\begin{document}

\maketitle
\thispagestyle{empty}
\pagestyle{empty}

\begin{abstract}
Deploying robots in real-world environments, such as households and manufacturing lines, requires generalization across novel task specifications without violating safety constraints.
Linear temporal logic (LTL) is a widely used task specification language with a compositional grammar that naturally induces commonalities among tasks while preserving safety guarantees.
However, most prior work on reinforcement learning with LTL specifications treats every new task independently, thus requiring large amounts of training data to generalize.
We propose LTL-Transfer, a zero-shot transfer algorithm that composes task-agnostic skills learned during training to safely satisfy a wide variety of novel LTL task specifications.
Experiments in Minecraft-inspired domains show that after training on only 50 tasks, LTL-Transfer can solve over 90\% of 100 challenging unseen tasks and 100\% of 300 commonly used novel tasks without violating any safety constraints.
We deployed LTL-Transfer at the task-planning level of a quadruped mobile manipulator to demonstrate its zero-shot transfer ability for fetch-and-deliver and navigation tasks.
\end{abstract}

\section{INTRODUCTION}
Deploying robots in the real world requires generalization across many novel tasks while preserving safety.
For example, an industrial robot that fetches the same components in different orders based on the assembled product should only learn to fetch each part once.
These tasks typically share constituents like objects and trajectory segments, which creates an opportunity to reuse knowledge~\cite{Taylor09}.

Linear temporal logic (LTL)~\cite{pnueli1977temporal} is an effective means of specifying objectives, including safety constraints, for reinforcement learning (RL) agents~\cite{littman2017environment, icarte2018lpopl, camacho2019ltl}.
Its compositional grammar reflects the compositional nature of most tasks.
However, most prior approaches to RL for LTL tasks learn to solve every new task from scratch.
We propose a zero-shot transfer algorithm, LTL-Transfer, that exploits the compositionality of LTL task specification to safely solve novel tasks without additional training by composing skills learned in prior tasks.
Transferring skills is more data-efficient than learning from scratch and more computationally efficient than planning.

\begin{figure}[!ht]
\centering
     \begin{subfigure}[b]{0.2\textwidth}
         \centering
         \includegraphics[width=\textwidth]{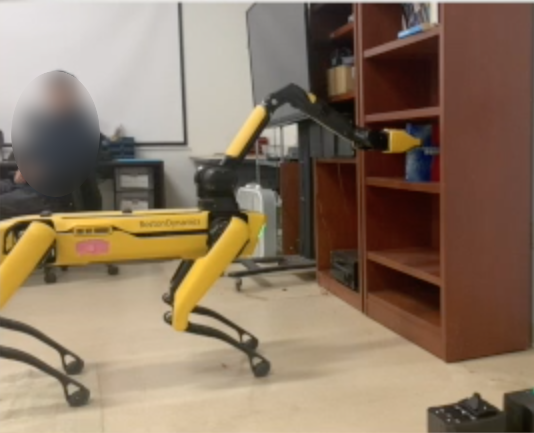}
         \caption{go to shelf to fetch book}
         \label{fig:robot_fig_fetch_book}
     \end{subfigure}
     \begin{subfigure}[b]{0.183\textwidth}
         \centering
         \includegraphics[width=\textwidth]{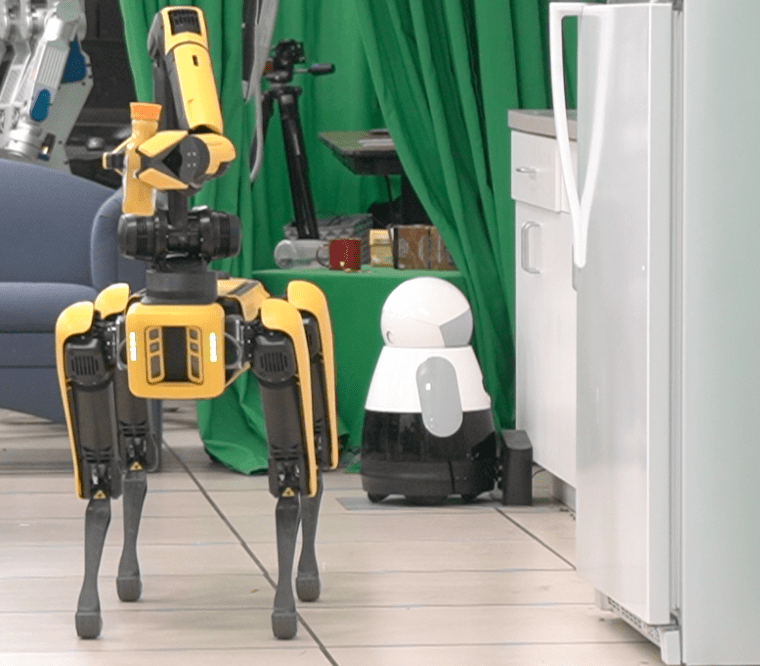}
         \caption{deliver juice to desk}
         \label{fig:robot_fig_fetch_juice}
     \end{subfigure}
     \caption{The robot is executing four task-agnostic skills sequentially to solve a novel task $\F (book \wedge \F (desk_a \wedge \F(juice \wedge \F desk_a)))$, i.e., fetch and deliver a book then a juice bottle to the user. Two of the four skills are shown.}
     \label{fig:robot}
\vspace{-10pt}
\end{figure}

We show in a simulated Minecraft-inspired domain that LTL-Transfer can solve over 90\% of 100 challenging unseen tasks and 100\% of 300 common and less complex novel tasks after training on only 50 tasks and never violates a safety constraint.
We deploy LTL-Transfer at the task-planning level of a quadruped mobile manipulator to solve fetch-and-deliver and navigation tasks in zero-shot.
Our key insight is efficiently reusing learned skills by leveraging similarities between the novel and training tasks.
Code, datasets, supplementary materials, and robot demonstration videos are at~\url{https://jasonxyliu.github.io/LTL-Transfer}.

\section{Preliminaries}\label{sec:pelim}
\textbf{Linear Temporal Logic (LTL) for Task Specification:} LTL is a widely used alternative to numerical rewards for task specification.
An LTL formula $\varphi$ is a Boolean function that determines whether a given trajectory satisfies the objective expressed by the formula.
Littman et al.~\cite{littman2017environment} showed that LTL can express non-Markovian, temporal tasks that numerical rewards cannot, and it has become a target language for acquiring task specification in many settings, including from natural language~\cite{liu23lang2ltl, patel20} and learning from demonstration~\cite{shah2018bayesian}.
Formally, an LTL formula is interpreted over traces of Boolean propositions over discrete time and is defined through the following recursive syntax:
\begin{equation}
    \varphi := \alpha \mid \neg \varphi \mid \varphi_1 \vee \varphi_2 \mid \X \varphi \mid \varphi_1 ~\U ~ \varphi_2.
\end{equation}

$\alpha\in \bm{\alpha}$ is a proposition that maps a state to a Boolean value; $\varphi$, $\varphi_1$, and $\varphi_2$ are valid LTL formulas.
The operator $\X$ (next) is used to define a property $\X \varphi$ that holds if $\varphi$ holds at the next time step.
The formula $\varphi_1~ \U ~\varphi_2$ with the binary operator $\U$ (until) specifies that $\varphi_1$ must hold until $\varphi_2$ first holds at a future time.
The operators $\neg$ (not) and $\vee$ (or) are identical to propositional logic operators.
We also utilize the following abbreviated operators: $\wedge$ (and), $\F$ (finally or eventually), and $\G$ (globally or always).
$\F \varphi$ specifies that the formula $\varphi$ must hold at least once in the future, and $\G \varphi$ specifies that $\varphi$ must always hold in the future. 
Consider the Minecraft map depicted in Figure \ref{fig:run_example}.
The task of collecting $wood$ and $axe$ in an arbitrary order is specified by the LTL formula $\F axe~ \wedge ~\F wood$. 
The formula $\F(axe ~\wedge~ \F wood)$ specifies collecting $wood$ at least once after collecting $axe$.
$\F wood ~\wedge ~ \neg wood~\U~ axe$ specifies the task of collecting $wood$ only after $axe$ has been collected.

Every LTL formula can be represented as a B\"uchi automaton~\cite{vardi1996automata, gerth1995simple} interpreted over an infinite trace of truth values of the propositions in the formula, thus providing an automated translation of an LTL specification to a transition-based representation.
We consider task specification in the co-safe fragment of LTL~\cite{kupferman2001model, pnueli1990hierarchy} where formulas can be verified by a finite trace, thus making it ideal for episodic tasks. 
Camacho et al.~\cite{camacho2019ltl} showed that every co-safe LTL formula $\varphi$ can be translated to an equivalent reward machine (RM)~\cite{icarte2022reward, icarte2018using}, $\mathcal{M}_\varphi = \tuple{ \mathcal{Q}_{\varphi}, q_{0, \varphi},\mathcal{Q}_{term, \varphi}, \varphi, T_{\varphi}, R_{\varphi}}$; where $\mathcal{Q}_\varphi$ is the finite set of states, $q_{0,\varphi}$ is the initial state, $\mathcal{Q}_{term, \varphi}$ is the set of terminal states; $T_\varphi: \mathcal{Q}_{\varphi} \times 2^{|\bm{\alpha}|} \rightarrow \mathcal{Q}_{\varphi}$ is the deterministic transition function; and $R_\varphi: \mathcal{Q_\varphi} \rightarrow \mathbb{R}$ represents the reward accumulated by entering a given state.
Figure~\ref{fig:run_example}d shows the reward machine graph representing the LTL formula $\F wood ~\wedge~ \neg wood ~ \U axe$.
Nodes encode progressed RM states, 0 for an accepting state and 3 for a failure state.
Boolean formulas label edges.
Truth assignments of propositions $\bm{\alpha}$ that satisfy edges induce transitions in the RM.
Our proposed algorithm, LTL-Transfer, for transferring learned policies to solve novel LTL specifications, is compatible with all algorithms that generate policies by solving a product MDP of the reward machine $\mathcal{M}_\varphi$ and the task environment.

\textbf{Options Framework:} Sutton and Barto \cite{sutton99option} introduced a framework, termed options or skills, for incorporating temporally extended actions into reinforcement learning. 
An option $o = \tuple{\mathcal{I}, \beta, \pi}$ is defined by the initiation set $\mathcal{I}$, a set of states where the option policy can be executed; the termination condition $\beta$, which determines when the execution ends; and the option policy $\pi$.
Our proposed algorithm, LTL-Transfer, compiles policies learned during training into task-agnostic options that are transferred to solve novel tasks.

\section{Related work}

Most approaches extending reinforcement learning (RL) to temporal tasks first generate a product MDP of the state space and the automaton equivalent of the LTL task specification~\cite{littman2017environment, icarte2018using, icarte2018lpopl, camacho2019ltl, icarte2022reward}.
Notably, Jothimurugan et al.~\cite{jothimurugan2021compositional} proposed interleaving graph-based planning on the automaton with RL to bias exploration towards trajectories that satisfy the LTL specification.
Although these approaches exploited the compositional structure of LTL to accelerate learning, they did not exploit the compositionality to transfer to novel tasks.
Thus, the policy to satisfy a novel LTL specification must be learned from scratch.

\begin{figure*}[th!]
    \centering
    \includegraphics[width=0.95\textwidth]{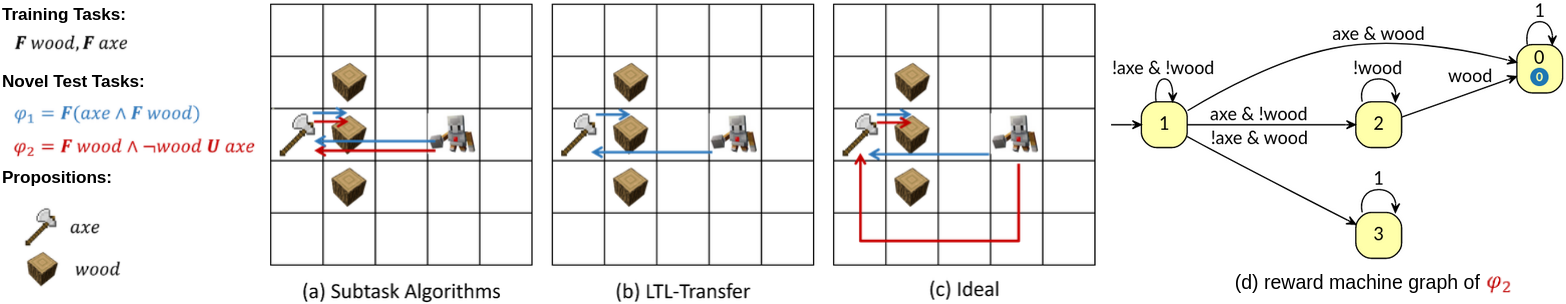}
    \caption{An example of tasks, the environment, and trajectories.
    The robot learned to solve the two training LTL tasks and is expected to solve two novel tasks $\varphi_1$ and $\varphi_2$. 
    Figure 2a depicts the trajectories output by a subtask-based algorithm (blue for $\varphi_1$, red for $\varphi_2$).
    Figure 2b depicts the trajectories produced by our proposed algorithm LTL-Transfer.
    Note that LTL-Transfer does not start execution for the task $\varphi_2$ as the two learned policies do not guarantee the preservation of the ordering constraint.
    Figure 2c depicts the optimal trajectories for the novel tasks $\varphi_1$ and $\varphi_2$.
    Figure 2d is a graph representation of the reward machine (RM) for the task $\varphi_2$.
    Nodes represent RM states.
    Edges represent Boolean formulas.
    }
    \label{fig:run_example}
    \vspace{-15pt}
\end{figure*}

A common approach towards generalization across temporal tasks has been to learn independent policies for subtasks~\cite{leon2020systematic, leon2021nutshell, araki2021lof, andreas2017modular}.
Given a new task, these methods then sequentially compose the learned policies in an admissible order. 
Consider the Minecraft-inspired grid world depicted in Figure~\ref{fig:run_example}a containing $wood$ and $axe$ objects.
The subtask-based methods learn policies to solve subtasks $\F wood$ and $\F axe$ involving reaching each object. 
When tasked with the specification $\varphi_{test} = \F wood~ \wedge~ (\neg wood ~\U ~axe)$, i.e., collect $wood$, but do not collect $wood$ until $axe$ is collected, the subtask-based methods would violate the ordering constraint by reaching $axe$ through the grid cells containing $wood$.
These approaches rely on additional fine-tuning to solve the target task correctly.
We propose a general framework for transferring learned policies to novel tasks in zero-shot without violating any safety constraints.

Kuo et al.~\cite{kuo2020encoding} proposed learning subnetworks for propositions and operators, then creating the final policy network through composition.
Vaezipoor et al.~\cite{vaezipoor2021ltl2action} proposed learning a latent embedding over LTL formulas using a graph neural network to solve novel LTL tasks.
In contrast, our method uses formal methods to identify learned policies best suited for transfer, thus requiring orders of magnitude fewer training tasks to achieve comparable results.
Furthermore, neither method can guarantee the preservation of safety constraints like our approach.
Xu et al.~\cite{xu2019transfer} considered transfer learning between pairs of source and target tasks, while our approach trains on a collection of tasks and transfers to a set of novel tasks.
Nangue Tasse et al.~\cite{tasse2022skill} attempted zero-shot composition through logical and temporal compositions of policies, but their approach assumes that a given task must be satisfiable.
By leveraging the structure of the task automaton, our proposed algorithm aborts execution immediately after it identifies the task as unsolvable given the learned skills.

A qualitatively distinct approach to zero-shot generalize to novel tasks is model-based RL that plans with an estimated transition model~\cite{sutton1998rl, moerland2023model}.
Compared to model-based methods, our approach transfers learned policies in zero-shot thus requires significantly less computation at test time.

Our approach draws inspiration from prior works on learning portable skills in Markov domains~\cite{konidaris2007building, james2020learning, bagaria2019option, bagaria2021robustly}.
These approaches learn task-agnostic representations of preconditions, constraints, and effects of an option~\cite{sutton99option}.
We learn portable skills to satisfy novel LTL task specifications.

\section{Problem definition}\label{sec:def}
We represent the environment as an MDP without the reward function $\mathcal{M}_\mathcal{S} = \langle \mathcal{S}, \mathcal{A}, T_{\mathcal{S}} \rangle$, where $\mathcal{S}$ is the set of states, $\mathcal{A}$ is the set of actions, and $T_{\mathcal{S}}: \mathcal{S} \times \mathcal{A} \times \mathcal{S} \rightarrow [0,1]$ represents the transition dynamics of the environment which we assume to be hidden from the RL algorithm.
Further, a set $\bm{\alpha}$ of Boolean propositions represents the facts about the environment and forms the compositional building blocks for specifying tasks.
We assume a labeling function $L: \mathcal{S} \rightarrow 2^{|\bm{\alpha}|}$ that maps the state to the Boolean propositions is given.
A task in the environment $\mathcal{M}_\mathcal{S}$ is specified by a linear temporal logic (LTL) formula $\varphi$, and it is translated to a reward machine $\mathcal{M}_\varphi = \tuple{ \mathcal{Q}_\varphi, q_{0,\varphi},\mathcal{Q}_{term, \varphi}, \varphi, T_\varphi, R_\varphi}$ detailed in Section~\ref{sec:pelim}.

Given a set of training tasks $\varPhi_{train} = \{ \varphi_1, \varphi_2, \ldots , \varphi_n\}$, specified by LTL, and policies for a set of options $\mathcal{O}_q$ learned from these training tasks, a zero-shot transfer algorithm needs to solve a novel LTL task $\varphi_{test} \not\in \varPhi_{train}$ in the same environment without additional training.

\section{LTL-Transfer with transition-centric options}\label{sec:methods}
\subsection{Algorithm Overview}\label{ss:overview}
Consider the environment map depicted in Figure~\ref{fig:run_example}b.
Assume an RL algorithm has learned option policies to collect $axe$ and $wood$ individually from the training tasks $\F axe$ and $\F wood$, respectively. 
Now given the test task $\varphi_1 = \F (axe ~\wedge ~ \F ~wood)$, i.e., first collect $axe$, then $wood$, a transfer algorithm should identify that sequentially composing the policies for $\F axe$ and $\F wood$ solves the new task $\varphi_1$ (as depicted in blue).
Then consider a different test task $\varphi_2 = \F wood ~\wedge~ \neg wood ~ \U axe$, i.e., first collect $axe$, then $wood$, but avoid $wood$ until $axe$ is collected.
The transfer algorithm must realize that the policy that satisfies $\F axe$ does not guarantee avoiding $wood$ while going to $axe$.
Therefore, it must not start execution using only these two learned policies to avoid accidentally violating the ordering constraint.

We developed a zero-shot transfer algorithm, LTL-Transfer, that composes learned policies to solve novel LTL tasks while enforcing ordering constraints.
It operates in two stages.
\begin{enumerate}
    \item First, LTL-Transfer accepts the set of training tasks $\varPhi_{train}$ and the policies learned from the training tasks and compiles a set of task-agnostic, portable options $\mathcal{O}_{e}$.
    \item Next, it identifies and executes a sequence of options from the set $\mathcal{O}_e$ to solve a novel task $\varphi_{test}$. 
\end{enumerate}

We can use a class of RL algorithms that operate on a product MDP composed of the environment $\mathcal{M}_\mathcal{S}$ and the reward machine $\mathcal{M}_\varphi$ to learn option policies from the training LTL tasks~\cite{littman2017environment, icarte2018using, icarte2018lpopl, camacho2019ltl, icarte2022reward}.
We chose LPOPL~\cite{icarte2018lpopl} because it explicitly allows for sharing policies across LTL task specifications that share progression states.
Given a set of LTL tasks, LPOPL first translates each task to a reward machine (RM) and decomposes tasks to subtasks, each of which corresponds to a state in the RM, then learns an action-value function, represented by a DQN~\cite{mnih2013dqn}, for each subtask.
Please see the supplementary materials for implementation details of LPOPL.
The learned policy is Markov with respect to the environment states $\mathcal{S}$ for a given RM state, i.e., the policy to be executed in the RM state $q \in \mathcal{Q}_\varphi$ is $\pi^\varphi_q:  \mathcal{S} \rightarrow \mathcal{A}$.

Executing the state-centric  option $o^\varphi_q\in\mathcal{O}_q$ with the policy $\pi^\varphi_q$ from the reward machine state $q$ triggers a transition in the RM on a path to an acceptance state.
There can be multiple outgoing transitions from an RM state, so the target RM transition of an option $o^\varphi_q$ is conditioned on the environment state $s\in\mathcal{S}$ where the execution was initiated.
We propose compiling each state-centric option, $o^\varphi_q$, into multiple transition-centric options by partitioning the initiation set of the state-centric option based on the estimated success probability of its policy satisfying the target RM transition from the starting environment state.
Each resulting transition-centric option will maintain the satisfaction of self-transition edge $e^\varphi_{q,q}$ from the starting RM state $q$ until it triggers the target RM transition $e^\varphi_{q,q'}$.
Our insight is that each transition-centric option triggers a transition in the reward machine on a path to an acceptance state, and these RM transitions may be shared across different tasks.
Thus, the transition-centric options $\mathcal{O}_e$ are portable across different tasks.
We describe the compilation algorithm in Section~\ref{ss:relabel}.

Given a novel LTL task specification $\varphi_{test} \not\in \varPhi_{train}$, our transfer algorithm first constructs a reward machine representing the task, $\mathcal{M}_{\varphi_{test}}$, then identifies a path through the reward machine that can be traversed by sequentially executing transition-centric options from the set $\mathcal{O}_e$. 
Our transfer algorithm is sound, and it terminates.
We describe the details of the transfer algorithm in Section~\ref{ss:transfer}.

The key advantage of our approach is that the option compilation can be done offline for any environment.
We can then transfer the options to solve novel tasks at execution time in zero-shot.
Thus, learning to solve a limited number of LTL tasks can help solve a wide gamut of unseen tasks.

\subsection{Compilation of Transition-Centric Options}\label{ss:relabel}
The policy learned to satisfy an LTL task specification $\varphi$ identifies the current reward machine state $q \in\mathcal{Q}_\varphi$ and executes a Markov policy $\pi^\varphi_q$ until the state of the reward machine progresses.
We represent this policy as a state-centric option, $o^\varphi_q = \tuple{\mathcal{S}, \beta_{e^\varphi_{q,q}}, \pi^\varphi_q}$, where the initiation set is the entire state space $\mathcal{S}$ of the environment; the option terminates when the truth assignment of the propositions $\bm{\alpha}$ violates the self-transition $e^\varphi_{q,q}$ from the RM state $q$.
The termination condition $\beta_{e^\varphi_{q,q}}$ is formally defined as follows,
\begin{equation}
    \beta_{e^\varphi_{q,q}} = \begin{cases}
    1, & \text{if } L(s) \nvDash e^\varphi_{q,q}\\
    0, & \text{otherwise}.
    \end{cases}
\end{equation}

A transition-centric option $o_{e^\varphi_{q,q}, e^\varphi_{q,q'}}$ executes a Markov policy that ensures that the truth assignment of the propositions $\bm{\alpha}$ satisfies the self-transition edge $e^\varphi_{q,q}$ at all time until the policy yields a truth assignment that satisfies the target outgoing edge $e^\varphi_{q,q'}$.
We define a transition-centric option as follows, 
\begin{equation}
    o_{e^\varphi_{q,q}, e^\varphi_{q,q'}} = \tuple {\mathcal{S}, \beta_{e^\varphi_{q,q}}, \pi^\varphi_q, e^\varphi_{q,q}, e^\varphi_{q,q'}, f_{e^\varphi_{q,q'}}}.
\end{equation}
The initiation set is the entire state space $\mathcal{S}$ of the environment; the termination condition $\beta_{e^\varphi_{q,q}}$ is defined by the violation of the self-transition edge $e^\varphi_{q,q}$; the option's policy is Markov $\pi^\varphi_q:\mathcal{S}\rightarrow \mathcal{A}$; $e^\varphi_{q,q}$ and $e^\varphi_{q,q'}$ represent the self-transition and the target outgoing edge, respectively; and $f_{e^\varphi_{q,q'}}: \mathcal{S} \rightarrow [0,1]$ represents the success probability of the policy $\pi^\varphi_q$ satisfying the target edge $e^\varphi_{q,q'}$ starting from the environment state $s\in \mathcal{S}$.

Algorithm \ref{alg:compile} describes our approach to compiling each state-centric option $o^\varphi_q$ to a set of transition-options $\set{o_{e^\varphi_{q,q},e^\varphi_{q,q'}}: q \in \mathcal{Q}_\varphi, q' \text{is out-neighbor of } q}$.
Executing the policy $\pi^{\varphi}_q$ of the state-centric option may satisfy different outgoing edges $e^{\varphi}_{q,q'}$ of the reward machine state $q$ depending on what environment state $s\in\mathcal{S}$ the execution was initiated. 
Thus, the success probability $f_{e^{\varphi}_{q,q'}}$ acts as a soft segmenter of the state space $\mathcal{S}$; it can be estimated by policy rollouts from all environment states in discrete domains or using sampling-based methods \cite{bagaria2019option, bagaria2021robustly} in continuous domains.

\begin{algorithm}[t]
\scriptsize
\caption{Compile State-Centric Options to Transition-Centric Options
}
\begin{algorithmic}[1]
\Function{Compile}{$\mathcal{M}_\mathcal{S}, \varPhi_{train}, \mathcal{O}_q$} 
\State $\mathcal{O}_e \gets \emptyset$
\For{$\varphi \in \varPhi_{train}$}
    \State $\mathcal{M}_\varphi \gets$ \Call{GenerateRM}{$\varphi$}
    \State $\mathcal{O}^\varphi_q \gets \set{o^{\varphi'}_q \in \mathcal{O}_q: \varphi' = \varphi}$ 
    \For{$o^{\varphi}_q = \tuple{\mathcal{S}, \beta^\varphi_{e_{q,q}}, \pi^\varphi_q}  \in \mathcal{O}^{\varphi}_q$}
        \State $\mathcal{Q}_{out}=\set{q': q' \text{ is an out-neighbor of } q}$
        \State $\forall  q'\in\mathcal{Q}_{out}: \mathcal{E} \gets \set{(e^\varphi_{q,q}, e^\varphi_{q,q'}): e^\varphi_{q,q} \text{ is the self edge}}$
        \For{$s \in \mathcal{S}$}
            \State Generate $N_r$ rollouts from $s$ using $\pi^{\varphi}_q$
            \State Record edge transition frequencies $n_s(e^\varphi_{q,q'}) ~\forall~(e^\varphi_{q,q}, e^\varphi_{q,q'})\in \mathcal{E}$
            \State $\forall ~ q' \in\mathcal{Q}_{out}: f_{e^\varphi_{q,q'}}(s) \gets \frac{n_s(e^\varphi_{q,q'})}{N_r}$ \label{lin:learn_f}
        \EndFor
        \State $\mathcal{O}^\varphi_{q,e} \gets \set{o_{e^{\varphi}_{q,q}, e^{\varphi}_{q,q'}} = \tuple{\mathcal{S}, \beta_{e^\varphi_{q,q}}, \pi^\varphi_q, e^\varphi_{q,q}, e^\varphi_{q,q'}, f{e^{\varphi}_{q,q'}}}}$ 
        \State $\mathcal{O}_e \gets \mathcal{O}_e \cup \mathcal{O}^\varphi_{q,e}$
    \EndFor
\EndFor
\State \Return{$\mathcal{O}_e$}
\EndFunction
\end{algorithmic}
\label{alg:compile}
\end{algorithm}

\subsection{Transferring to Novel LTL Task Specification} \label{ss:transfer}
Algorithm~\ref{alg:transfer} describes the zero-shot transfer algorithm that composes transition-centric options from the set $\mathcal{O}_e$ to solve a novel test task $\varphi_{test}$ in the environment $\mathcal{M}_\mathcal{S}$.
Line~\ref{lin:generate} generates the reward machine (RM) graph for the test task.
Line~\ref{lin:prune} examines each edge $e^{\varphi_{test}}_{q,q'}$ of the RM, identifies the transition-centric options that satisfy that edge transition, and prunes an edge if no such option is found.
Line~\ref{lin:id_paths} identifies and caches all paths from the current state $q$ to the accepting state $q^{\top}$ of the reward machine.
Lines~\ref{lin:id_options1} and \ref{lin:id_options2} identify a set of all eligible options that can potentially achieve an RM transition from the current state to a progressed state on a path to the accepting state. 

Lines~\ref{lin:option} and \ref{lin:execute} then execute the option with the highest success probability of satisfying the target edge transition, estimated by $f$. 
If the option fails to progress to another RM state, we delete it from the set (Line~\ref{lin:delete}) and execute the next option.
If the set of eligible options is empty at any point without reaching the accepting RM state $q^{\top}$, Line~\ref{lin:fail} exits with a failure.
A successful transfer occurs when the RM progresses to the accepting state $q^{\top}$.

\begin{algorithm}[t]
\scriptsize
\caption{Zero-shot transfer to test task $\varphi_{test}$}
\begin{algorithmic}[1]
\Function{Transfer}{$\mathcal{M}_\mathcal{S}$, $\varphi_{test}$, $\mathcal{O}_e$}
\State $\mathcal{M}_{\varphi_{test}} \gets$ \Call{Generate\_RM}{$\varphi_{test}$} \label{lin:generate}
\State $\mathcal{M}_{\varphi_{test}} \gets \Call{Prune}{\mathcal{M}_{\varphi_{test}}}$ \label{lin:prune}
\State $s \gets$ \Call{Initialize}{$\mathcal{M}_\mathcal{S}$}
\State $q \gets q_{0,\varphi_{test}}$
\While{$q\neq q^{\top}$} 
    \State $P_{\text{cache}} \gets \set{p_i:~ p_i = [e_0,\ldots ,e_{n_i}] ~\text{path from } q \text{ to } q^{\top} \text{ in } \mathcal{M}_{\varphi_{test}}} $ \label{lin:id_paths}
    \State $\mathcal{O}_{p[0]}=\set{o_{e_1, e_2}\in\mathcal{O}_e: \Call{MatchEdge}{(e_1, e_2),(e^{\varphi_{test}}_{q,q}, p[0])}}$ $\forall p \in P_{\text{cache}}$ \label{lin:id_options1} 
    \State $\mathcal{O}_{[0]} = \bigcup_p\mathcal{O}_{p[0]}$ \label{lin:id_options2}
    \State $\tuple{s',q'} \gets \tuple{s,q}$
    \While{$\mathcal{O}_{[0]} \neq \emptyset$ and $q' = q$}
        \State $o^*_{e_1,e_2} \gets \argmax_{o_{e_1, e_2}\in \mathcal{O}_{[0]}} f_{e_2}(s)$ \label{lin:option} 
        \State $\tuple{s',q'} \gets$ \Call{Execute}{$\pi^*$} \label{lin:execute}
        \If{$q' = q$}
        \State $\mathcal{O}_{[0]} \gets \mathcal{O}_{[0]}\setminus o^*_{e_1,e_2}$ \label{lin:delete} 
        \EndIf
    \EndWhile
    \If{$q' = q$}
        \State \Return Failure \label{lin:fail}
    \Else
        \State $\tuple{s,q} \gets \tuple{s',q'}$
    \EndIf
\EndWhile
\State \Return Success
\EndFunction
\end{algorithmic}
\label{alg:transfer}
\end{algorithm}

\subsection{Matching Options to Reward Machine Transitions}\label{ss:match}
The edge-matching conditions determine whether we can safely apply a transition-centric option to transition along an edge of the reward machine (RM).
We used the edge-matching conditions to prune the RM graph to retain only the edges matched with available options (Algorithm~\ref{alg:transfer} Line~\ref{lin:prune}) and identify eligible options from a given RM state (Algorithm~\ref{alg:transfer} Line~\ref{lin:id_options1}).
Given a test task $\varphi_{test}$, when the current RM state is $q$, its self-transition edge is $e^{\varphi_{test}}_{q,q}$ and the target outgoing edge is $e^{\varphi_{test}}_{q,q'}$, to determine if a transition-centric option $o_{e_1, e_2}$ matches the target transition in the RM, we propose two edge-matching conditions, \textit{Constrained} and \textit{Relaxed}.
Both ensure the task execution does not fail due to an unsafe transition.

\textbf{Constrained Edge-Matching Condition}
requires that every truth assignment satisfying the self-transition edge $e_1$ of the option also satisfies the self-transition edge $e^{\varphi_{test}}_{q,q}$ of the reward machine $\mathcal{M}_{\varphi_{test}}$. 
Similarly, every truth assignment satisfying the outgoing edge $e_2$ of the option must satisfy the target transition $e^{\varphi_{test}}_{q,q'}$ of the test task's RM.
This strict requirement reduces the applicability of the learned options but ensures that the target edge is always achieved.
The \textit{Constrained} edge-matching condition is satisfied if the following Boolean expression holds:

\begin{equation}
\neg sat(e_1 \wedge \neg e^{\varphi_{test}}_{q,q} )~ \wedge ~\neg sat(e_2 \wedge \neg e^{\varphi_{test}}_{q,q'}).
\end{equation}
Let $sat(g)$ be a Boolean function that is true if and only if a truth assignment exists to satisfy the Boolean formula $g$.

\textbf{Relaxed Edge-Matching Condition} requires the self edges $e_1$ and $e^{\varphi_{test}}_{q,q}$ share satisfying truth assignments, so must the outgoing edges $e_2$ and $e^{\varphi_{test}}_{q,q'}$.
However, it allows the option to have valid truth assignments that may not satisfy the target outgoing edge, yet none of the truth assignments should trigger a transition to an unrecoverable failure state $q^{\bot}$.
We identify $q^{\bot}$ as a sink state of the RM graph without outgoing edges.
Further, all truth assignments that terminate the option must not satisfy the self-transition edge of the test task's reward machine.
The \textit{Relaxed} edge-matching condition can retrieve more eligible options.
It is satisfied if the following Boolean expression holds:
\begin{multline}
    sat(e_1 \wedge e^{\varphi_{test}}_{q,q})~ \wedge ~sat(e_2 \wedge e^{\varphi_{test}}_{q,q'})~ \wedge\\ 
    \neg sat(e_1 \wedge e^\bot) ~\wedge~ \neg sat(e_2 \wedge e^\bot)~ \wedge~ \neg sat(e_2 \wedge e^{\varphi_{test}}_{q,q}).
\end{multline}

\subsection{Optimization}
We parallelized the two key computational bottlenecks of LTL-Transfer, i.e., the estimation of the success probability $f_{e^{\varphi}_{q,q'}}$ and the evaluation of the edge-matching condition since there are no shared memory requirements.
We implemented the \textit{Relaxed} edge-matching condition using a propositional model counting algorithm~\cite{valiant1979complexity} from SymPy~\cite{sympy}, and the \textit{Constrained} edge-matching condition using string comparison to circumvent the model counting problem and found a significant speed-up.

\section{Experiments}\label{exp}
The aim of our evaluation is to test the hypothesis that LTL-Transfer can efficiently transfer learned skills to solve novel temporal tasks while preserving safety guarantees.
We tested our hypothesis in simulation and on a physical robot. 
The simulation domain allows us to scale skill transfer across a wide variety of tasks, while the real robot demonstrates the practicality of our approach for real-world mobile manipulation tasks.
We first defined five types of specifications of varying complexity for training and testing, then conducted experiments to evaluate the following hypotheses,




 \begin{enumerate}
    \item \textbf{H1:} LTL-Transfer exceeds the baselines' success rates of solving novel tasks.
    \item \textbf{H2:} \textit{Relaxed} edge-matching condition leads to higher success rates than the \textit{Constrained} condition.
    \item \textbf{H3:} Transferring learned policies to certain specification types (introduced in Section~\ref{ss:order_type}) leads to higher success rates.
    \item \textbf{H4:} LTL-Transfer is robust to highly uncertain transition dynamics.
\end{enumerate}

\subsection{Task Environment}\label{ss:task_env}
We tested LTL-Transfer in a Minecraft-inspired grid-world domain
commonly seen in compositional reinforcement learning (RL) and integration of temporal logics with RL~\cite{andreas2017modular, icarte2018lpopl, jothimurugan2021compositional, araki2021lof}.
This domain is particularly well suited for testing transfer learning with temporal tasks as policies to solve individual tasks can be learned rapidly with few computational resources.
Thus, we can run comprehensive evaluations of skill transfer across a full factorial of training and test task types that cover a wide variety of LTL templates.
We conducted experiments on four maps, similar to that depicted in Figure~\ref{fig:run_example}, with a dimension of $19\times19$.
Each location in the map is either vacant or occupied by one of nine object types.
Multiple instances of an object type may occur across the map.
After the agent enters a grid cell occupied by an object, the proposition representing that object type becomes true.
Actions are moving in four cardinal directions; an invalid action results in no movement. 
We tested zero-shot transfer in both deterministic and stochastic domains.

\begin{figure*}[t]
\centering
     \begin{subfigure}[b]{0.195\textwidth}
         \centering
         \includegraphics[width=\textwidth]{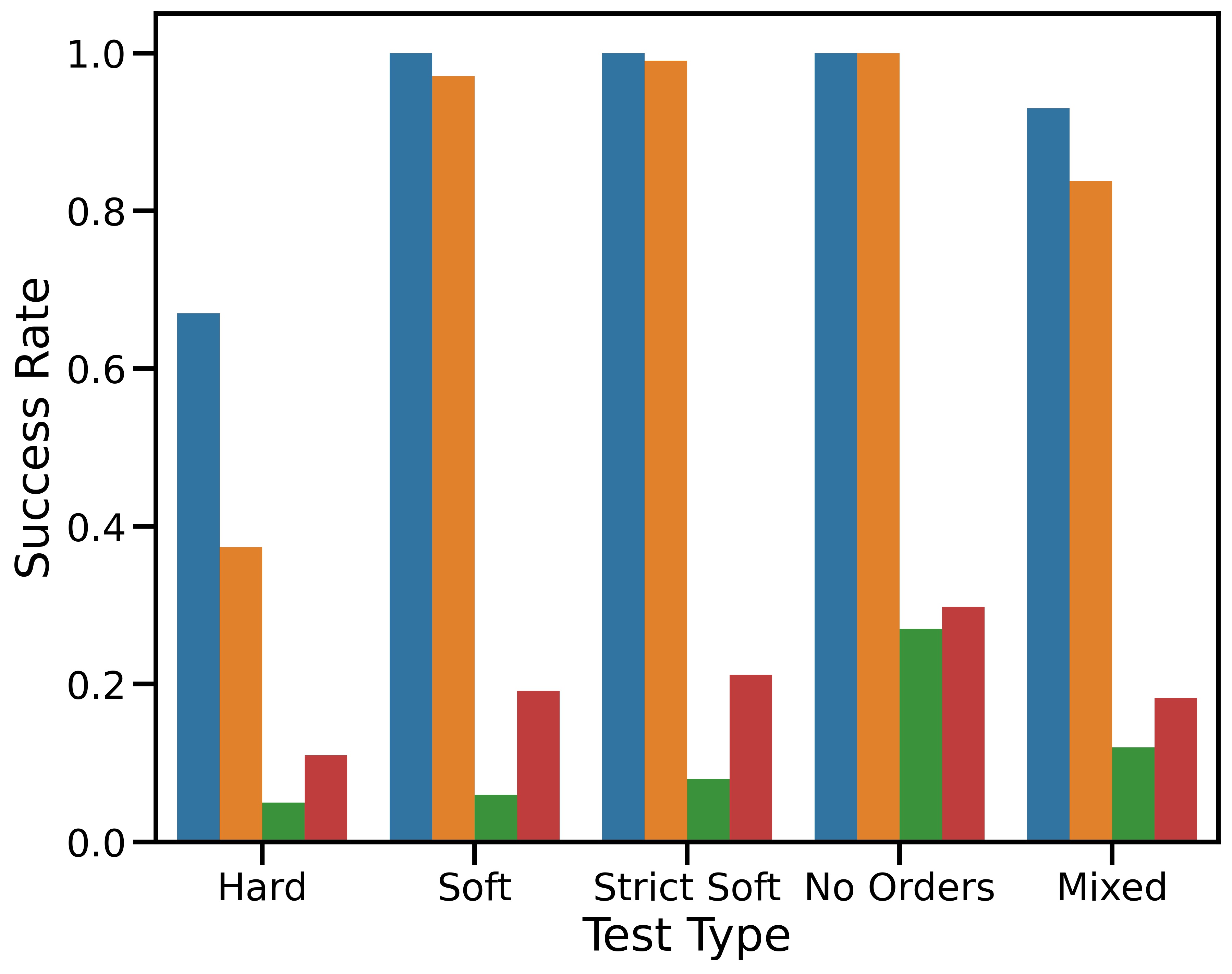}
         \caption{Success Rate}
         \label{fig:baseline_success}
     \end{subfigure}
    \begin{subfigure}[b]{0.195\textwidth}
         \centering
         \includegraphics[width=\textwidth]{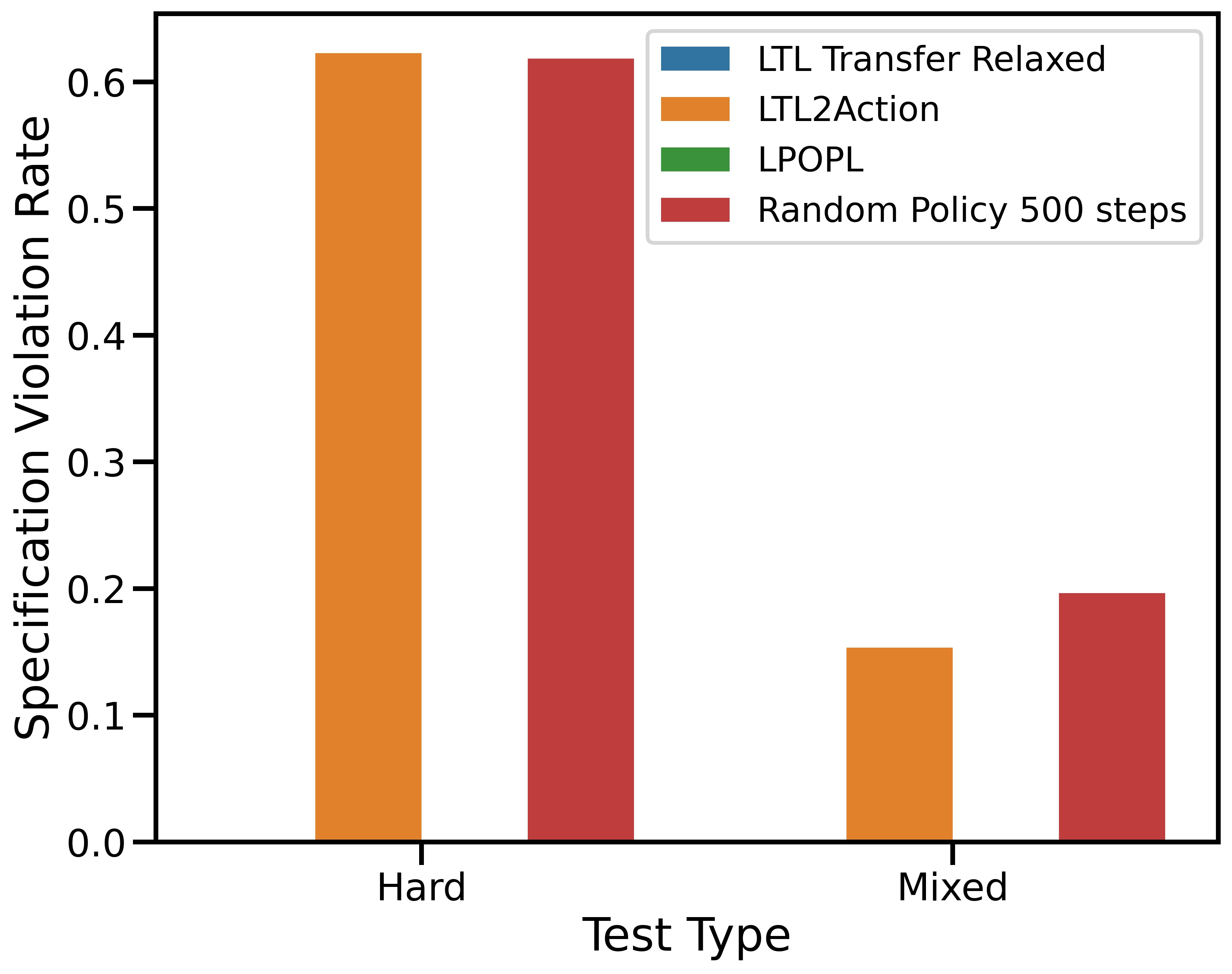}
         \caption{Violation Rate}
         \label{fig:baseline_violation}
    \end{subfigure} 
    \begin{subfigure}[b]{0.195\textwidth}
         \centering
         \includegraphics[width=\textwidth]{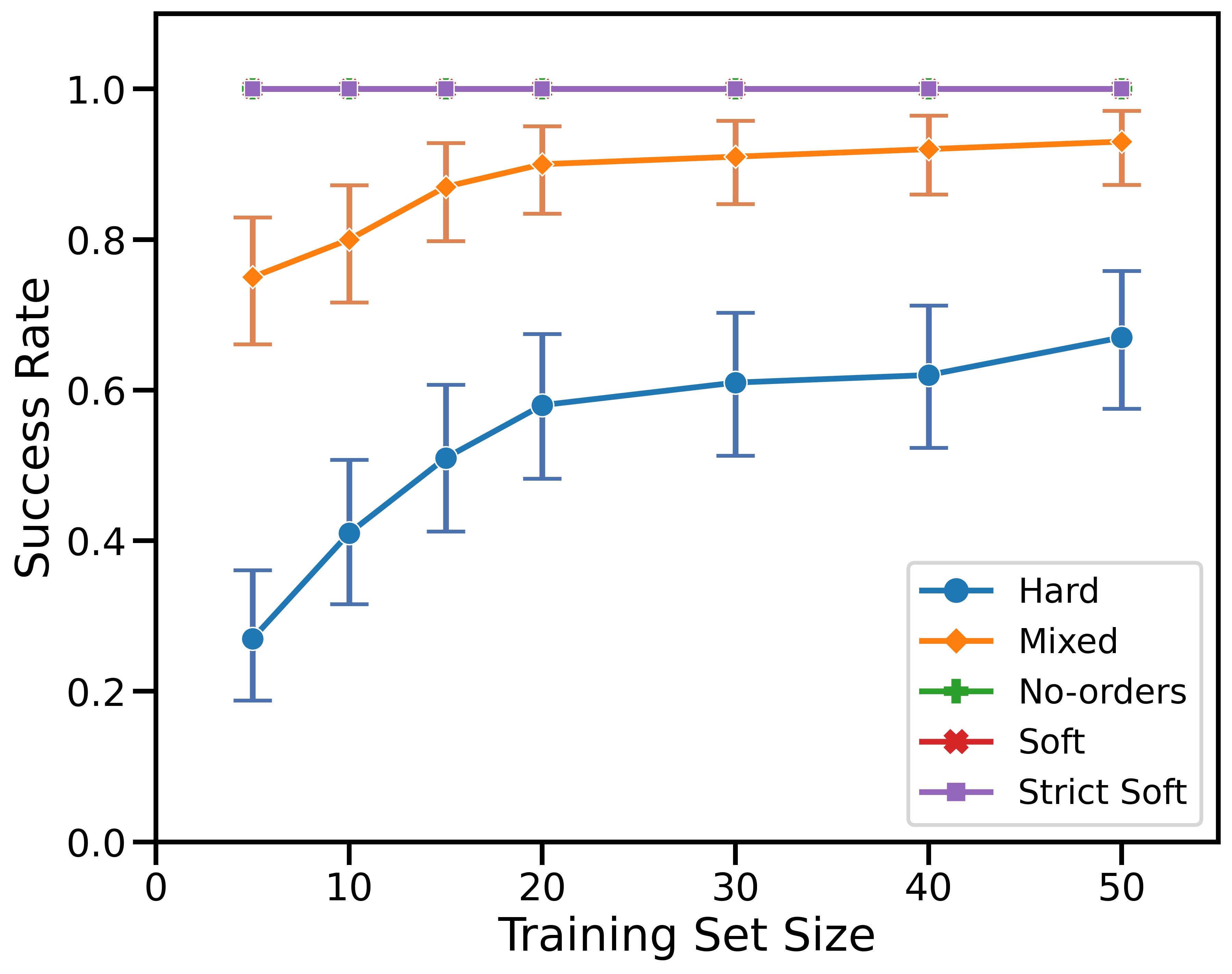}
         \caption{\textit{Relaxed} Match}
         \label{fig:relaxed}
     \end{subfigure}
     \begin{subfigure}[b]{0.195\textwidth}
         \centering
         \includegraphics[width=\textwidth]{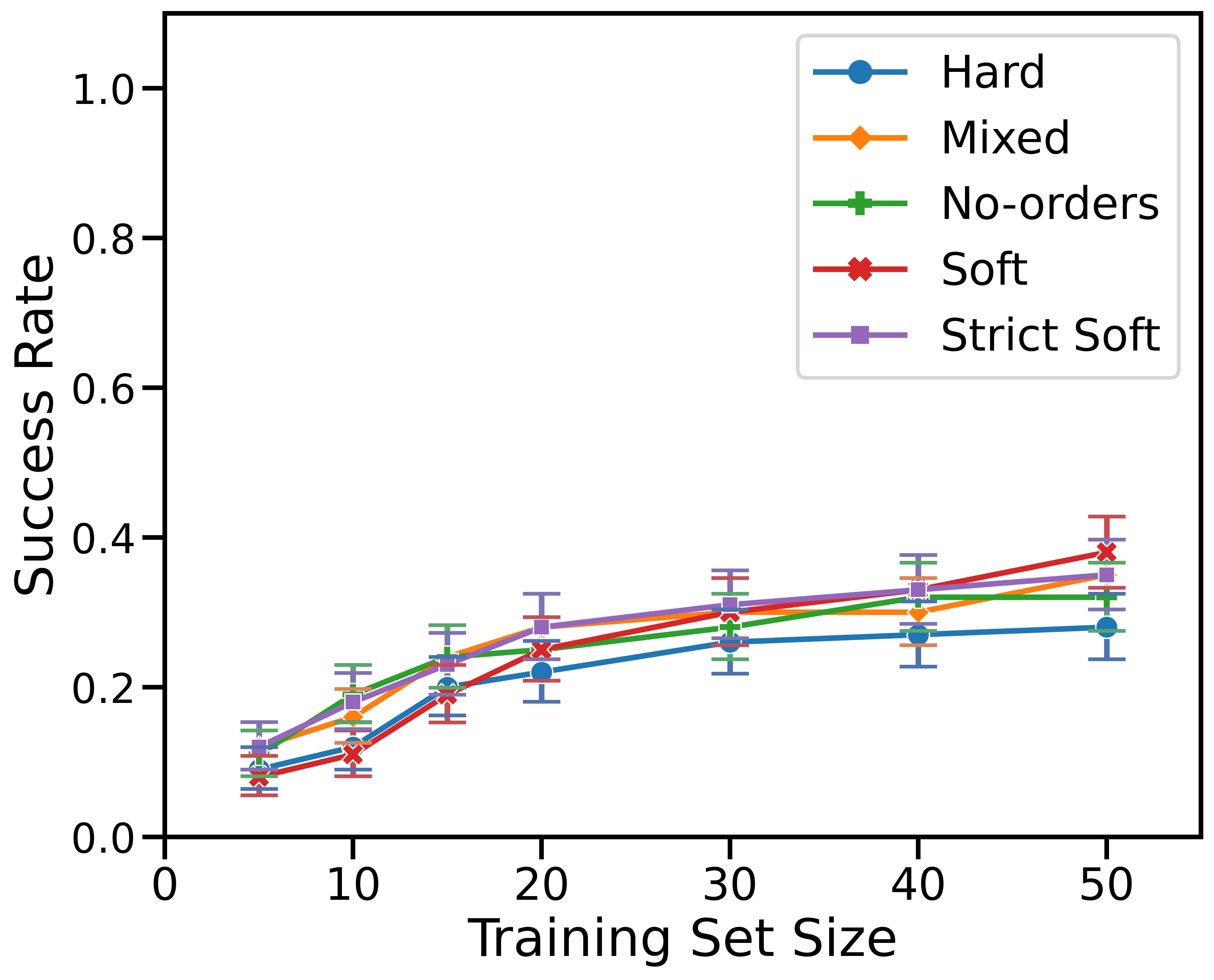}
         \caption{\textit{Constrained} Match}
         \label{fig:constrained}
     \end{subfigure}
     \begin{subfigure}[b]{0.195\textwidth}
         \centering
         \includegraphics[width=\textwidth]{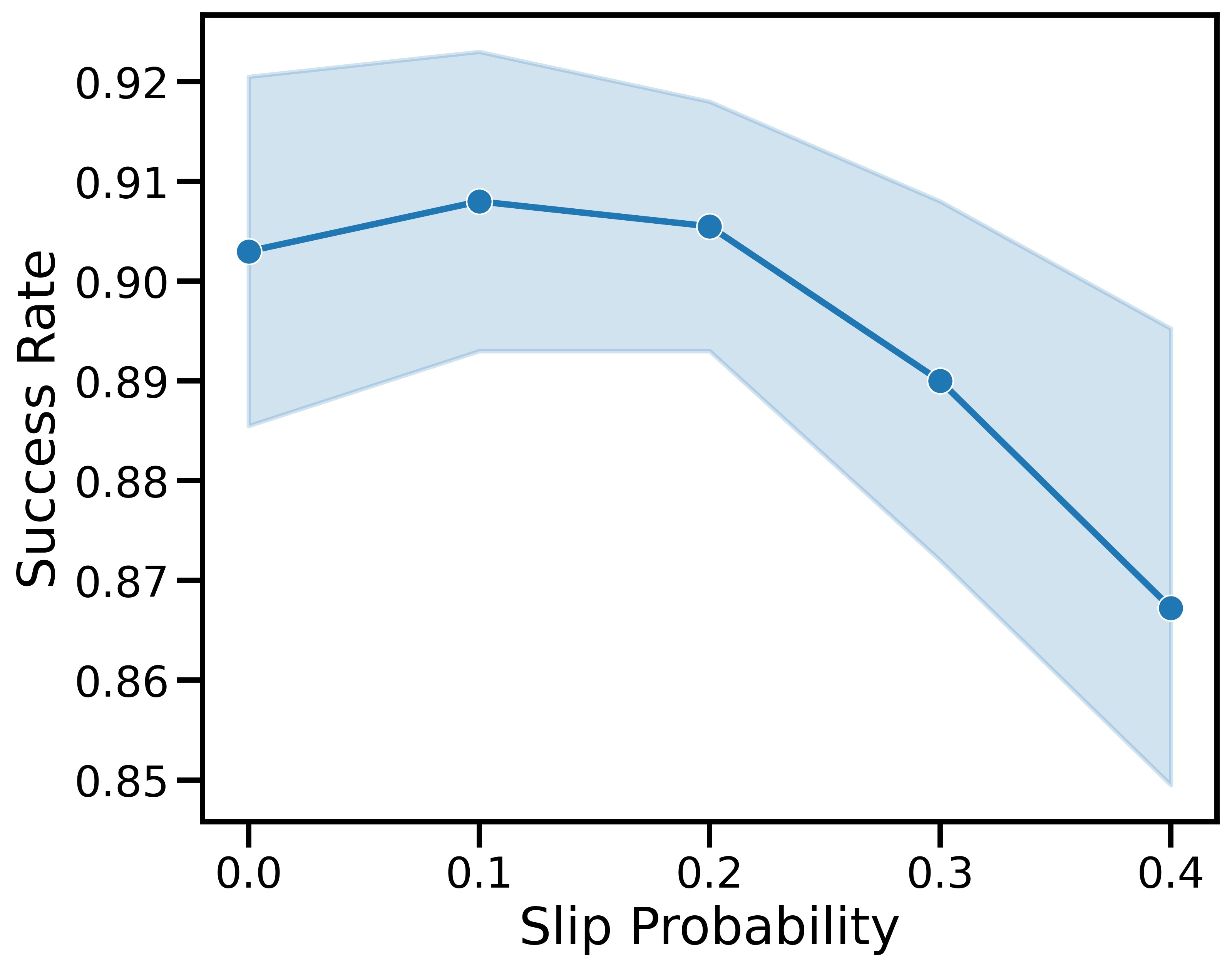}
         \caption{Stochastic Domain}
         \label{fig:stochastic}
     \end{subfigure}
\caption{
Figure~\ref{fig:baseline_success} shows the success rates of four methods on five test sets after training on the \textit{Mixed} training set.
Figure~\ref{fig:baseline_violation} depicts their specification violation rates (with a shared legend).
Figure~\ref{fig:relaxed} and~\ref{fig:constrained} show the success rates of LTL-Transfer on five test sets after training on \textit{Mixed} sets of various sizes using \textit{Relaxed} and \textit{Constrained} edge-matching condition, respectively.
The error bars depict the 95\% credible interval if the successful transfer was modeled as a Bernoulli distribution.
Figure~\ref{fig:stochastic} shows the success rate as the slip probability increases averaged over four maps.
}
\label{fig:1}
\vspace{-10pt}
\end{figure*}

\subsection{Types of Task Specifications}
\label{ss:order_type}
For a comprehensive evaluation of transferring learned policies across various LTL specifications, we considered the following three types of ordering constraints, proposed by Shah et al.~\cite{shah2018bayesian}, which constitute five specification types.
Each constraint is defined on a pair of propositions $a$ and $b$.
Without loss of generality, we assume that $a$ precedes $b$.

\begin{enumerate}
    \item \textbf{Hard} ordering constraints occur when $b$ must never be true before $a$.
    This property is expressed through the LTL formula $\neg b ~\U ~a$.
    \item \textbf{Soft} ordering constraints allow $b$ to occur before $a$ as long as $b$ happens at least once after $a$ becomes true for the first time. Soft orders are expressed through the LTL formula $\F(a ~\wedge~ \F b)$.
    \item \textbf{Strictly Soft} ordering constraints are similar to soft orders except that $b$ must be true strictly after $a$ first holds. Thus, $a$ and $b$ holding simultaneously would not satisfy a strictly soft order.
    Strictly soft orders are expressed through the LTL formula $\F(a ~\wedge~ \X\F b)$.
\end{enumerate}

We sampled five training sets, with 50 formulas in each, that represent five different specification types, \textit{Hard}, \textit{Soft}, \textit{Strictly-Soft}, \textit{No-Orders}, and \textit{Mixed}, and a test set of 100 formulas for each type.
This imitates the real-world scenario where users do not know the test task beforehand, so the robot must be trained on a limited set of training tasks then transfer to a wide variety of novel tasks.
When constructing a test set, we excluded tasks already in the training set.
All specifications in the \textit{Hard}, \textit{Soft}, and \textit{Strictly-Soft} sets were expressed using the respective templates described above.
\textit{No-Orders} sets have no ordering constraints, e.g., $\F a~\wedge~\F b~\wedge~\F c$.
In the $\textit{Mixed}$ set, each binary precedence constraint was expressed as one of the three ordering constraints described above. 
A given task in the environment involves visiting a specified set of object types in an admissible order determined by ordering constraints.
Please see the supplementary materials for example formulas from each specification type.

\subsection{Results and Discussion}

\textbf{Comparison with Baselines:}
We compare the performance of LTL-Transfer to three baselines, i.e., LTL2Action \cite{vaezipoor2021ltl2action}, LPOPL \cite{icarte2018lpopl}, and a random policy.

\textbf{LTL2Action} proposed by Vaezipoor et al. \cite{vaezipoor2021ltl2action} embeds LTL specifications using a graph neural network and sequentially selects the next proposition to satisfy.
This pre-trained embedding is appended to the state features to yield a goal-conditioned task policy, termed LTL Bootcamp, which serves as the upper bound of LTL2Action's performance on novel tasks.
We trained the LTL Bootcamp on the same training sets as LTL-Transfer and compared their performance.

\textbf{LPOPL}, detailed in Section~\ref{ss:overview}, serves as the lower bound that any transfer algorithm must surpass due to its limited transfer ability to solve novel tasks.
While LPOPL was not explicitly designed for zero-shot transfer, it can satisfy task specifications that are a progression of a training LTL formula because of its use of LTL progression and multi-task learning.




Figure~\ref{fig:baseline_success} depicts the success rate of all the methods on 100 novel tasks in each test set described in Section~\ref{ss:order_type} after training on 50 tasks of the \textit{Mixed} type.
LPOPL performed worse than the random policy as it did not attempt to satisfy any specification that did not exist in its progression set.
Both LTL-Transfer and LTL2Action have near-perfect success rates on \textit{Soft}, \textit{Strictly-Soft}, and \textit{No-Orders} test set as these specifications do not have irrecoverable failure state.
Specifications in \textit{Hard} and \textit{Mixed} test sets have failure states.
Thus, we observe a lower success rate across all methods.
However, LTL-Transfer demonstrates the best transfer success rate in the difficult test sets.
Crucially, Figure~\ref{fig:baseline_violation} shows that LTL-Transfer never violated any specification, while LTL2Action's specification violation rate is similar to that of the random policy, which could mean that LTL2Action essentially acts randomly given an infeasible task.



\textbf{Effect of Edge-Matching Condition:} We trained LTL-Transfer on \textit{Mixed} training sets of varying sizes and tested zero-shot transfer on all five test sets.
The success rates of using the \textit{Relaxed} and \textit{Constrained} edge-matching conditions are depicted in Figure~\ref{fig:relaxed} and Figure~\ref{fig:constrained}, respectively.
We observed that LTL-Transfer using the \textit{Relaxed} edge-matching condition successfully transferred to significantly more novel specifications across all types, thus supporting \textbf{H2}.

\textbf{Relative Difficulty of Specification Types:}
Figure~\ref{fig:relaxed} shows that LTL-Transfer with the \textit{Relaxed} edge-matching condition can perfectly transfer to novel tasks of \textit{Soft}, \textit{Strictly-Soft} and \textit{No-Orders} types after training on very few tasks.
After training on 50 tasks, LTL-Transfer can transfer to over 95\% of novel tasks of the \textit{Mixed} type.
Tasks of the \textit{Hard} type are the most difficult to transfer to.
Figure~\ref{fig:constrained} shows that the different tasks are equally difficult to transfer to using the \textit{Constrained} edge-matching condition.
Thus, \textbf{H3} is supported only by using the \textit{Relaxed} edge-matching condition.

\textbf{Stochastic Environments:}
To evaluate the robustness of LTL-Transfer to stochastic transitions, we increased the slip probability from 0 to 0.4 and observed a slight 3\% decrease in transfer success in Figure~\ref{fig:stochastic}, which supports \textbf{H4}.


\subsection{Robot Demonstrations}
We deployed LTL-Transfer at the task-planning level of a quadruped mobile manipulator, Spot~\cite{spot}, with off-the-shelf motion planning and grasping capabilities in a discretized indoor environment where the robot can fetch and deliver objects while navigating through the space.
LTL-Transfer first learned policies from 20 training tasks, then transferred the learned skills to 100 novel tasks from the five specification types defined in Section~\ref{ss:order_type}, 50 of which were executed on the robot~\footnote{Videos are at~\url{https://jasonxyliu.github.io/LTL-Transfer}}.
Please see the supplementary materials for all test tasks executed on the robot.
The environment contains 31 grid cells, but LTL-Transfer works in larger domains, like $19\times19$, as presented in Section~\ref{ss:task_env}.
The state space includes the locations of the robot and six landmarks, i.e., two desks, a couch, a door, a bookshelf with a book, and a kitchen counter with a juice bottle on top.
Actions are moving in the four cardinal directions.
Only navigation skills are learned from the training tasks.
Picking is executed when the goal location of a navigation option is the bookshelf or the counter.
Placement is executed after the robot stops at the desk or the couch and has picked up an object.

\section{CONCLUSIONS}
We introduced LTL-Transfer, a zero-shot transfer algorithm that uses the compositionality of LTL task specification to maximally transfer learned policies to solve various novel tasks.
Experiments in deterministic and stochastic Minecraft-inspired domains showed that LTL-Transfer can solve complex unseen tasks without violating any safety constraints.
We deployed LTL-Transfer at the task-planning level of a mobile manipulator to safely solve fetch-and-deliver and navigation tasks in zero-shot.
We envision incorporating long-term planning and intra-option policy updates to produce not just satisfying but optimal solutions to novel tasks.




\section*{ACKNOWLEDGMENT}
The authors thank Peilin Yu for editing the robot demonstration videos.
This work is supported by ONR under grant numbers N00014-21-1-2584, N00014-17-1-2699, and N00014-22-1-2592, NSF under grant number CNS-2038897, Amazon Robotics under award number 1061079, and funding from Echo Labs.
Partial funding for this work was provided by The Boston Dynamics AI Institute (``The AI Institute'').


\clearpage
\bibliographystyle{IEEEtran}
\bibliography{references}

\clearpage
\begin{appendices}
\setcounter{table}{0}
\setcounter{figure}{0}

\section{The Implementation Details of LPOPL}
LPOPL~\cite{icarte2018lpopl}'s DQN policy consists of a feedforward network with two hidden layers, each of which has 64 ReLU units.
We used the same hyperparameters suggested in~\cite{icarte2018lpopl} for training, i.e., the learning rate is 0.0001; the size of the replay buffer is 25,000; 32 transitions are randomly sampled from the replay buffer for every update; the discount factor is 0.9; exploration decreases linearly from 1 to 0.02.


\section{Example LTL Formulas}
We provide LTL task specifications and their interpretations from the \textit{Hard}, \textit{Soft}, \textit{Strictly Soft}, \textit{No Orders}, and \textit{Mixed} formula types.

\textbf{Hard:} Example formulas and their interpretations from the \textit{Hard} type are as follows:
\begin{enumerate}
    \item $\F workbench ~ \wedge \F factory ~ \wedge ~ \F iron ~ \wedge ~ \F shelter ~ \wedge\\ ~ \neg factory ~\U ~axe$: Visit $workbench$, $factory$, $iron$, $shelter$, and $axe$.
    Ensure that $factory$ is not visited before $axe$.

    \item $\F toolshed ~ \wedge ~ \F bridge ~ \wedge ~ \F factory ~ \wedge ~ \F axe ~ \wedge\\ ~\neg bridge ~ \U ~ wood$: Visit $toolshed$, $bridge$, $factory$, $axe$, and $wood$.
    Ensure that $bridge$ is not visited before $wood$.

    \item $\F wood  ~ \wedge ~ \F axe ~ \wedge \neg  wood ~\U~ grass ~ \wedge ~ \\ \neg grass ~ \U~ workbench ~ \wedge ~ \neg workbench~ \U~ bridge $: Visit $bridge$, $workbench$, $grass$, $wood$, and $axe$.
    Ensure visiting $bridge$, $workbench$, $grass$, and $wood$ in that particular order.
    Objects that occur later in the sequence cannot be visited before any prior objects.    
\end{enumerate}

\textbf{Soft:} Example formulas and their interpretations from the $\textit{Soft}$ type are as follows:
\begin{enumerate}
    \item $\F (bridge ~\wedge ~ \F (factory ~ \wedge ~ \F (iron ~ \wedge ~ \F shelter)))$: Visit $bridge$, $factory$, $iron$, and $shelter$ in that sequence.
    The objects that occur later in the sequence may be visited before the prior objects, provided that they are visited at least once after the prior object has been visited.

    \item $\F workbench ~ \wedge \F ( factory ~ \wedge ~ \F grass)$: Visit the $workbench$, $factory$, and $grass$. Visit $grass$ at least once after visiting the $factory$.

    \item $\F (axe ~ \wedge ~ \F factory) ~ \wedge \F workbench$: Visit $axe$, $factory$, and $workbench$. Ensure that $factory$ is visited at least once after $axe$.
\end{enumerate}

\textbf{Strictly Soft:} Example formulas and their interpretations from the $\textit{Strictly Soft}$ type are identical to the \textit{Soft} specifications, except they do not allow simultaneous satisfaction of multiple subtasks.
The subtasks in the sequence must occur strictly after the prior subtask.
This is enforced using nested operators next and finally $\X \F a$ instead of $\F a$.

\textbf{No Orders:} These specifications only contain a list of subtasks to be completed.
No temporal orders are enforced between any two subtasks.
\begin{enumerate}
    \item $\F wood  ~ \wedge ~ \F grass ~ \wedge \F stone$: Collect $wood$, $grass$, $stone$ in no particular order.
\end{enumerate}

\textbf{Mixed:} Example formulas and their interpretations from the $\textit{Mixed}$ type are as follows:
\begin{enumerate}
    \item $\F toolshed ~ \wedge \F factory   ~ \wedge \neg toolshed ~\U ~shelter ~ \wedge \F (grass~ \wedge ~ \F bridge) $: Visit the $toolshed$, $factory$, $shelter$, $grass$, and $bridge$.
    Ensure that $toolshed$ is not visited before the $shelter$ and $bridge$ is visited at least once after $grass$.

    \item $\F grass  ~ \wedge ~ \neg grass ~\U ~ toolshed ~\wedge ~\\ \F (factory ~ \wedge ~ \X\F workbench)$: Visit $grass$, $toolshed$, $factory$, and $workbench$.
    Ensure that $grass$ is not visited before $toolshed$ and $workbench$ is visited at least once strictly after $factory$.

    \item $\F iron  ~ \wedge ~ \neg iron ~\U ~ toolshed ~\wedge ~ \F (shelter ~ \wedge ~ \X\F wood)$: Visit $iron$, $toolshed$, $shelter$, and $wood$.
    Ensure that $iron$ is not visited before $toolshed$ and $wood$ is visited at least once strictly after $shelter$.
\end{enumerate}

\section{Additional Experimental Results}
\textbf{Learning Curves for Various Training Sets:} We present learning curves of success rates for transferring policies learned on different specification types.

\begin{figure}[h!]
\centering
     \begin{subfigure}[b]{0.23\textwidth}
         \centering
         \includegraphics[width=\textwidth]{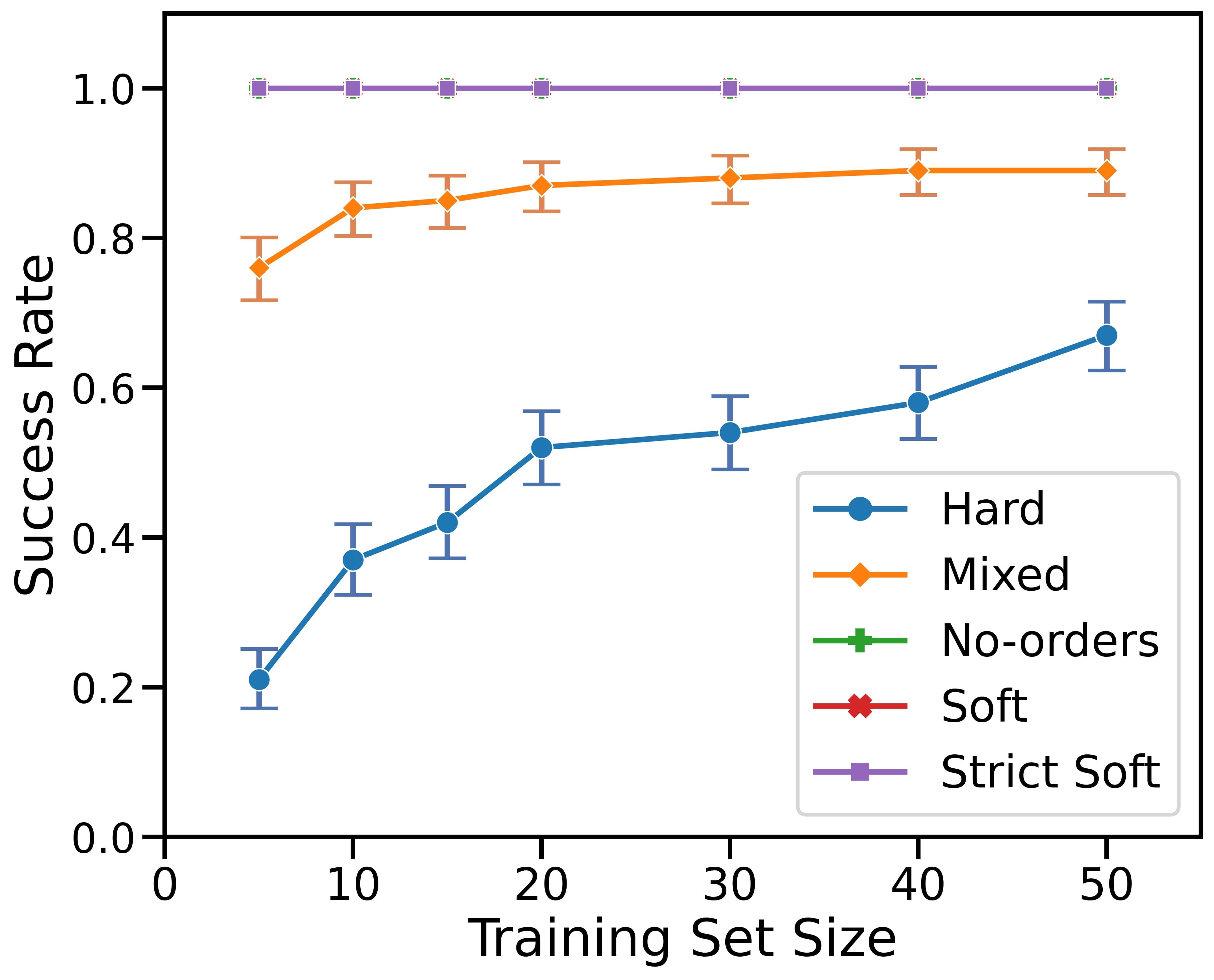}
         \caption{\textit{Relaxed} Edge Match}
         \label{fig:lc_relaxed_hard}
     \end{subfigure}
     \begin{subfigure}[b]{0.23\textwidth}
         \centering
         \includegraphics[width=\textwidth]{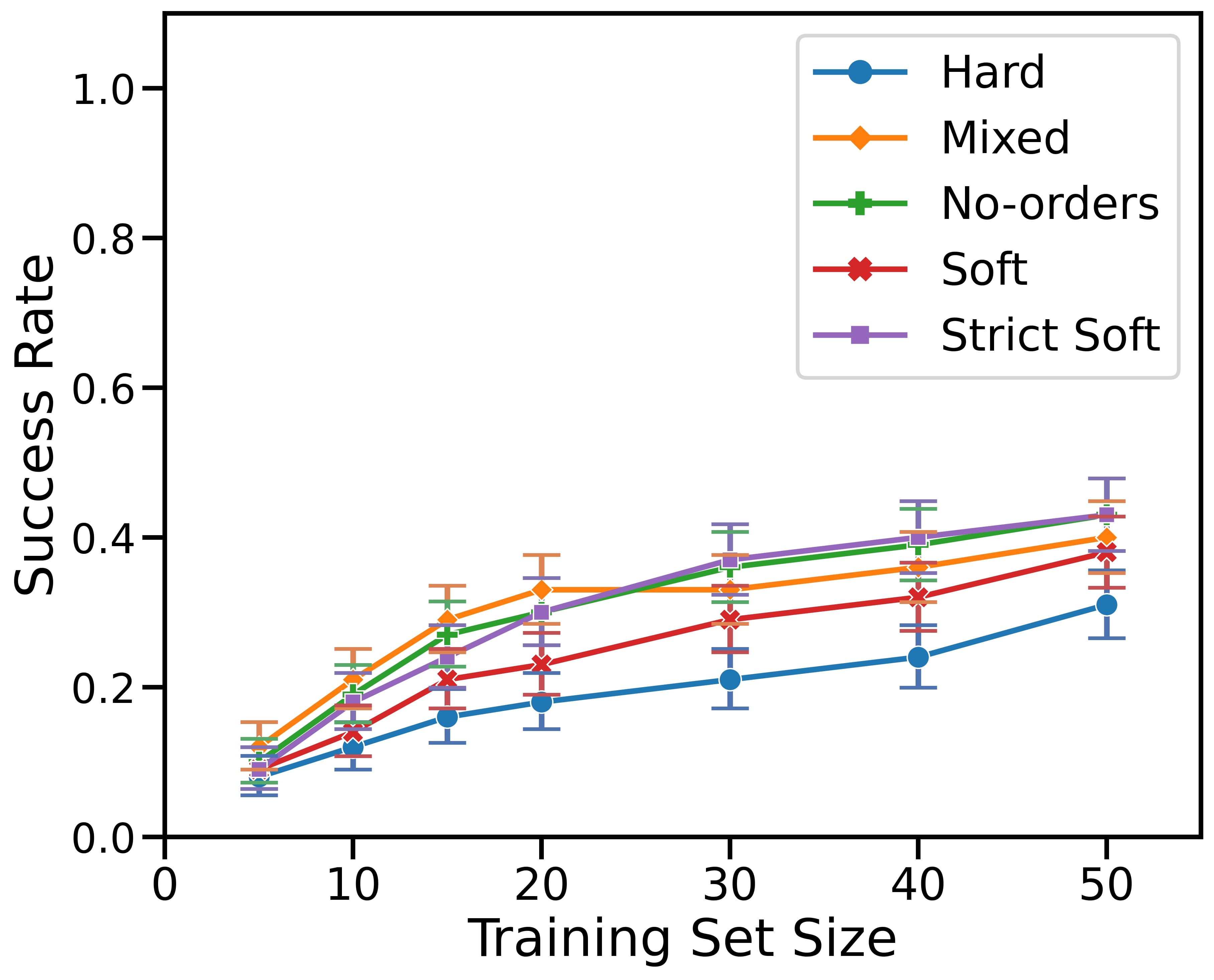}
         \caption{\textit{Constrained} Edge Match}
         \label{fig:lc_constrained_hard}
     \end{subfigure}
     \caption{
     Figure~\ref{fig:lc_relaxed_hard} depicts the success rates of transferring to five different specifications types using the~\textit{Relaxed} edge-matching condition after LTL-Transfer being trained on~\textit{Hard} training sets of various sizes.
     Figure~\ref{fig:lc_constrained_hard} depicts the success rates with the~\textit{Constrained} edge-matching condition. 
     Note that the error bars depict the 95\% credible interval if the successful transfer was modeled as a Bernoulli distribution.
     }
     \label{fig:hard}
\end{figure}

The learning curves for training on LTL tasks from the~\textit{Hard} training set with both~\textit{Relaxed} and~\textit{Constrained} edge-matching conditions are depicted in Figure~\ref{fig:hard}.

\begin{figure}[h!]
\centering
    \begin{subfigure}[b]{0.23\textwidth}
         \centering
         \includegraphics[width=\textwidth]{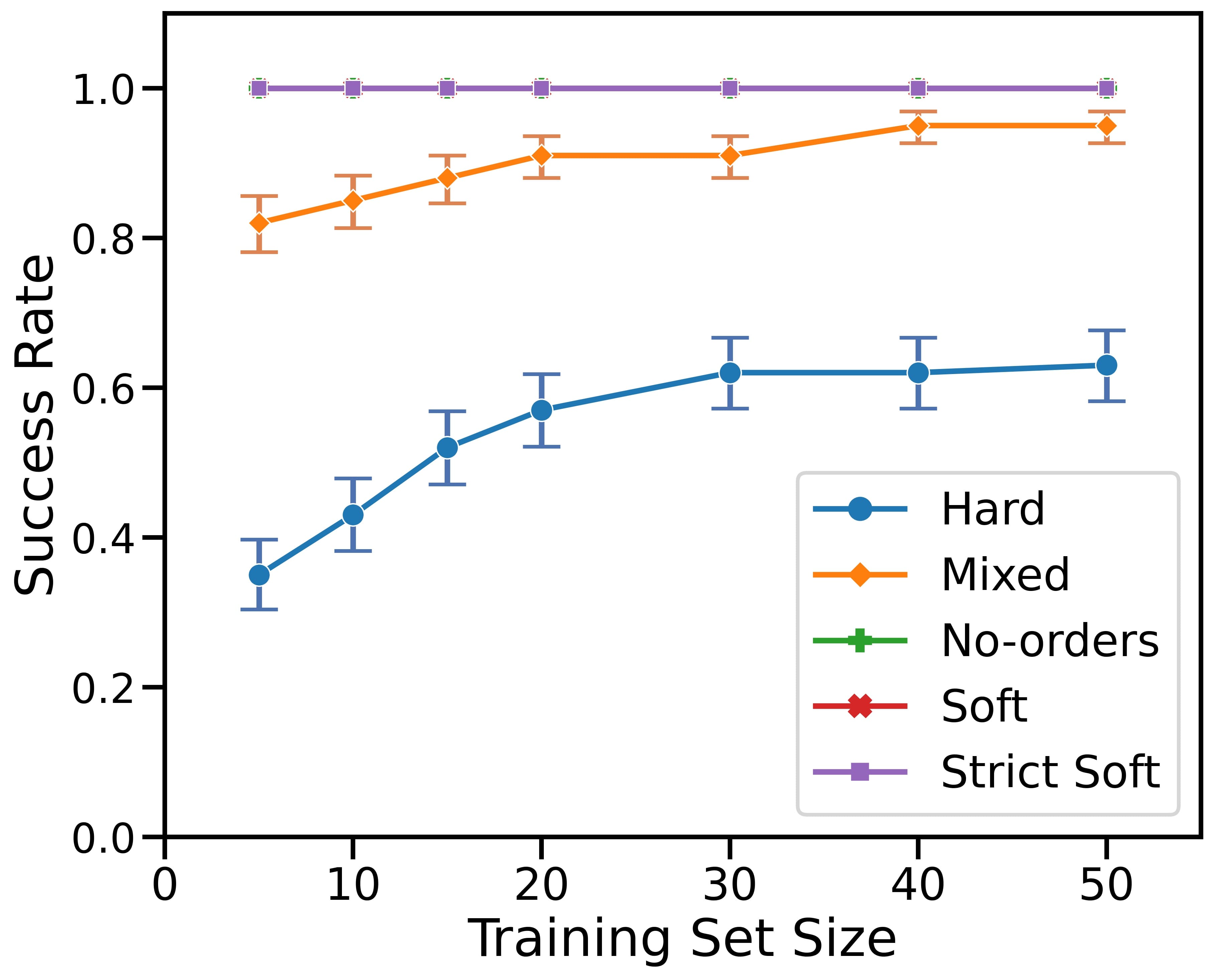}
         \caption{\textit{Relaxed} Edge Match}
         \label{fig:lc_relaxed_soft}
     \end{subfigure}
     \begin{subfigure}[b]{0.23\textwidth}
         \centering
         \includegraphics[width=\textwidth]{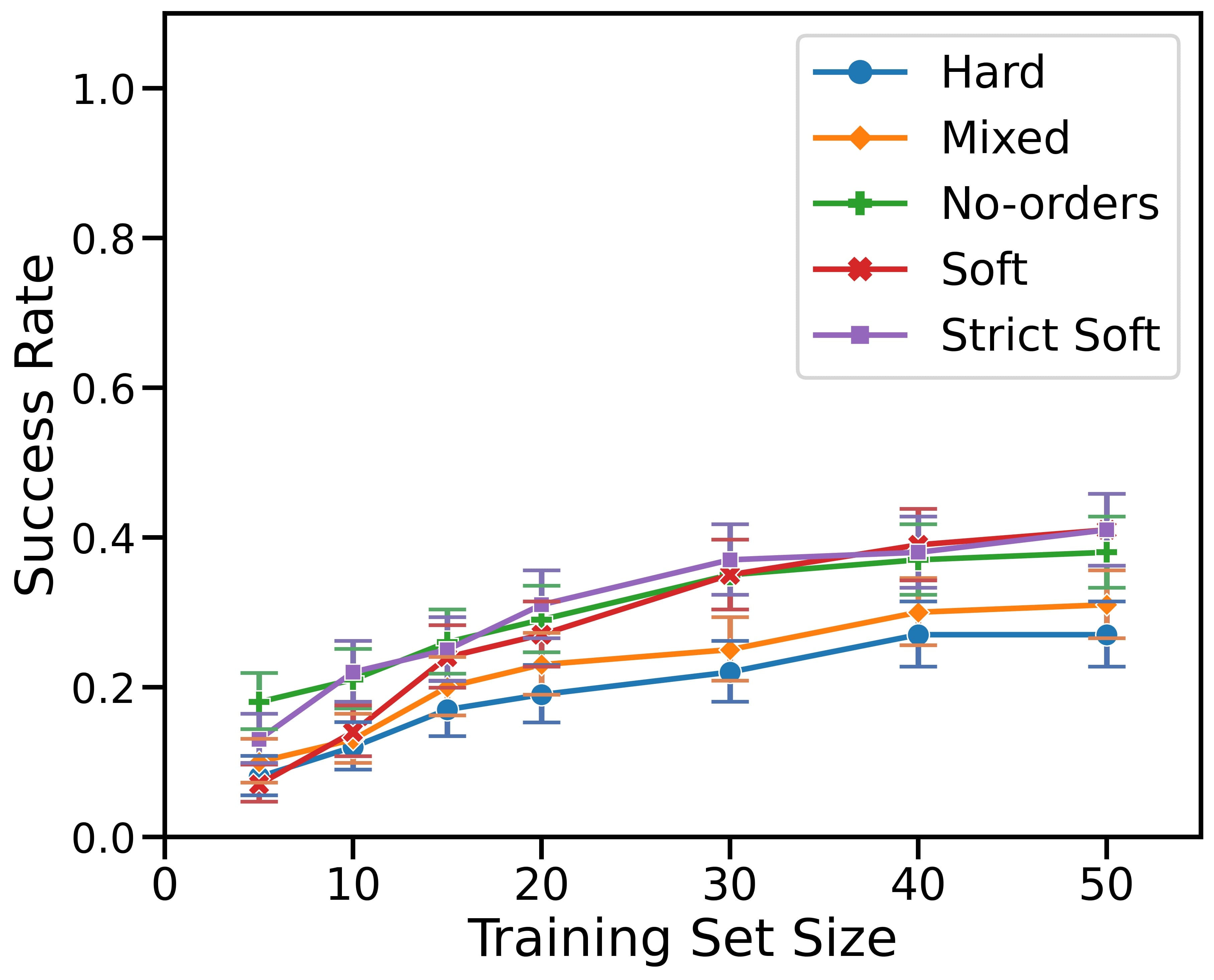}
         \caption{\textit{Constrained} Edge Match}
         \label{fig:lc_constrained_soft}
     \end{subfigure}
     \caption{
     Figure~\ref{fig:lc_relaxed_soft} depicts the success rates of transferring to five different specifications types using the~\textit{Relaxed} edge-matching condition after LTL-Transfer being trained on~\textit{Soft} training sets of various sizes.
     Figure~\ref{fig:lc_constrained_soft} depicts the success rates with the~\textit{Constrained} edge-matching condition.
     Note that the error bars depict the 95\% credible interval if the successful transfer was modeled as a Bernoulli distribution.
     }
     \label{fig:soft}
\end{figure}

The learning curves for training on LTL tasks from the~\textit{Soft} training set with both~\textit{Relaxed} and~\textit{Constrained} edge-matching conditions are depicted in Figure~\ref{fig:soft}.

\begin{figure}[h!]
\centering
    \begin{subfigure}[b]{0.23\textwidth}
         \centering
         \includegraphics[width=\textwidth]{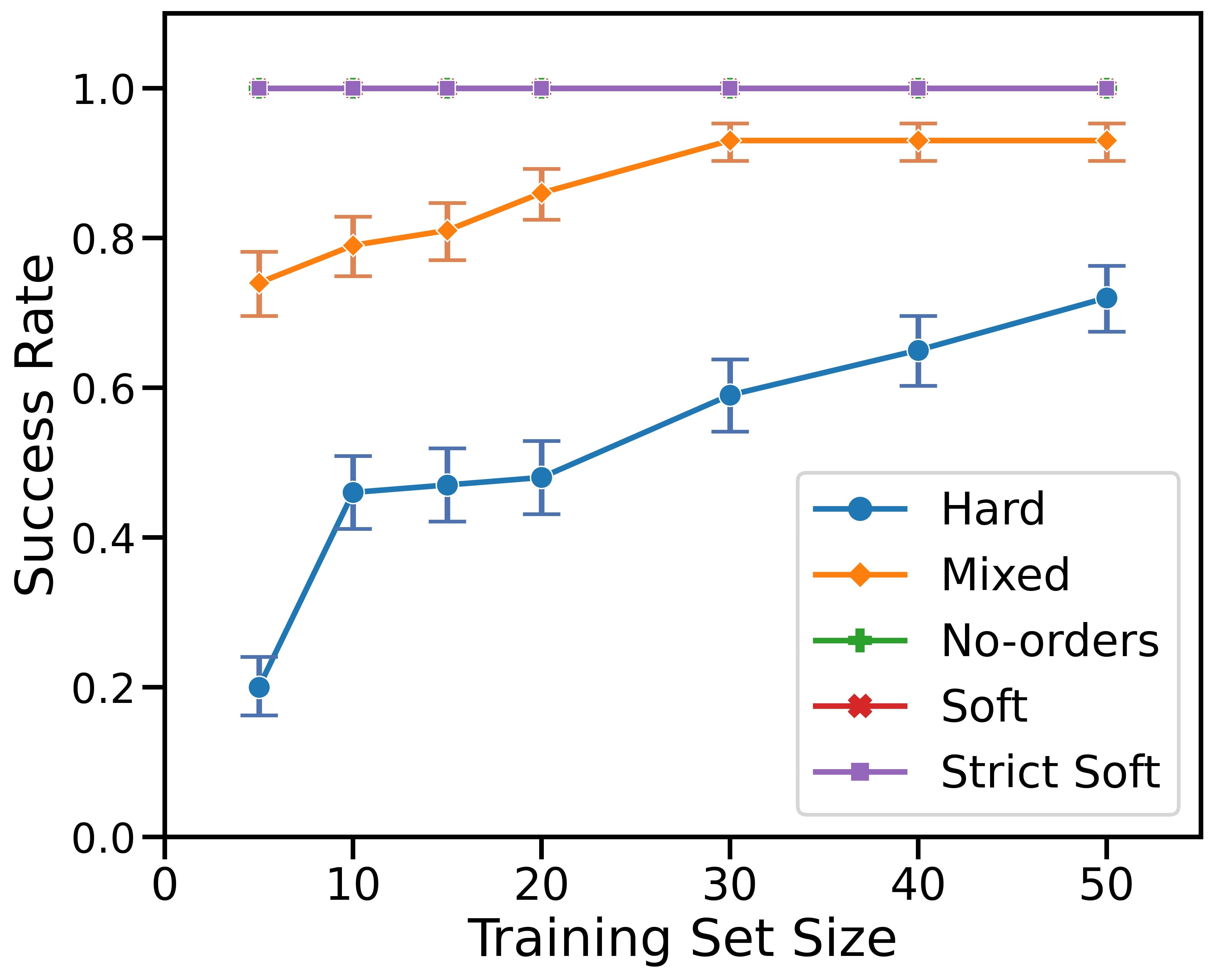}
         \caption{\textit{Relaxed} Edge Match}
         \label{fig:lc_relaxed_strict_soft}
     \end{subfigure}
     \begin{subfigure}[b]{0.23\textwidth}
         \centering
         \includegraphics[width=\textwidth]{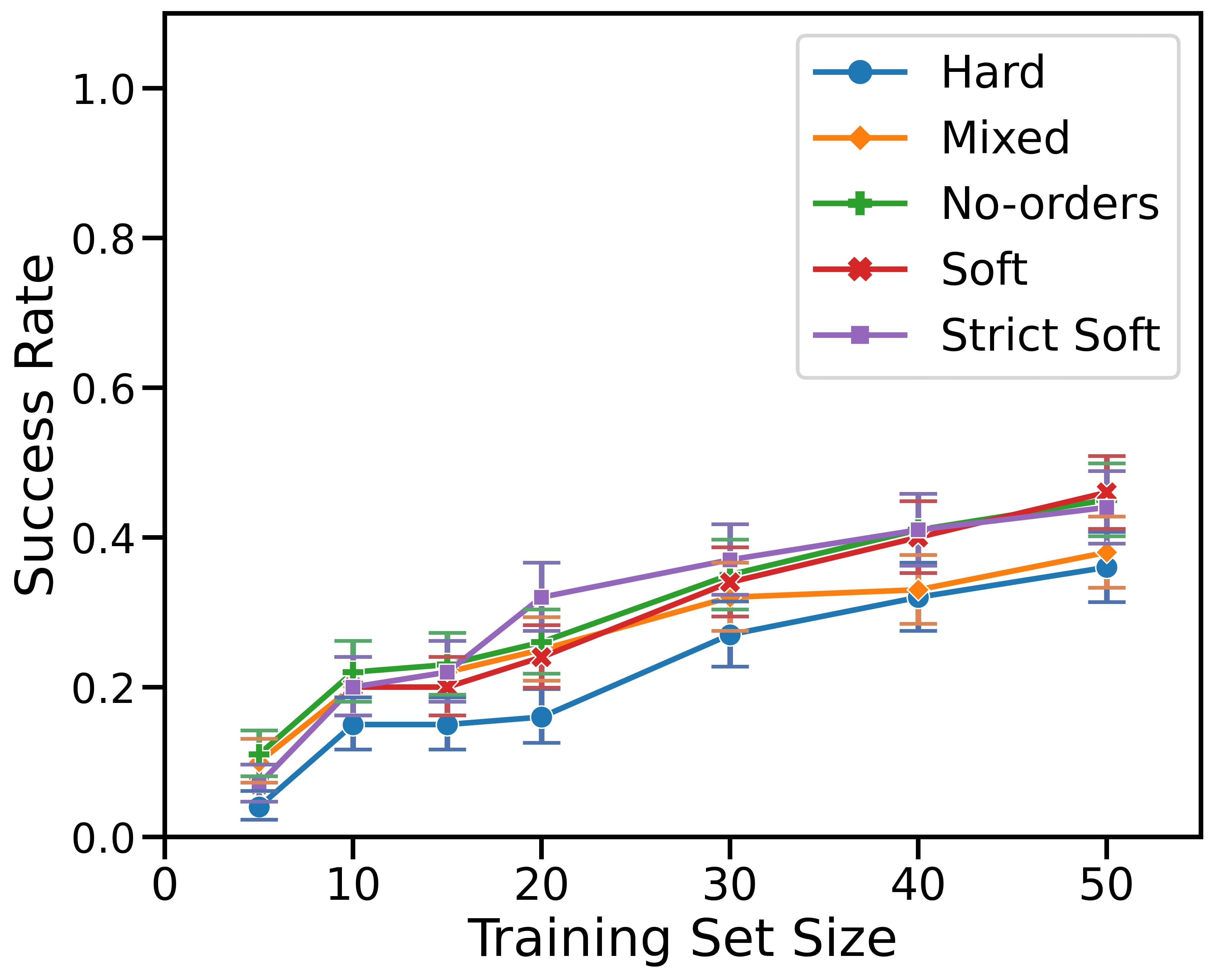}
         \caption{\textit{Constrained} Edge Match}
         \label{fig:lc_constrained_strict_soft}
     \end{subfigure}
     \caption{
     Figure~\ref{fig:lc_relaxed_strict_soft} depicts the success rates of transferring to five different specifications types using the~\textit{Relaxed} edge-matching condition after LTL-Transfer being trained on~\textit{Strictly Soft} training sets of various sizes.
     Figure~\ref{fig:lc_constrained_strict_soft} depicts the success rates with the~\textit{Constrained} edge-matching condition.
     Note that the error bars depict the 95\% credible interval if the successful transfer was modeled as a Bernoulli distribution.
     }
     \label{fig:strict_soft}
\end{figure}

The learning curves for training on LTL tasks from the \textit{Strictly Soft} training set with both~\textit{Relaxed} and~\textit{Constrained} edge-matching conditions are depicted in Figure~\ref{fig:strict_soft}.

The learning curves for training on LTL tasks from the \textit{No Orders} training set are expected to share nearly identical trends as the learning curves from the other training sets.

Note that for training on each specification type, the learning curve trends are nearly identical to the learning curves of training on the~\textit{Mixed} specification types, as depicted in Figure 3 in the main paper.
\textit{Hard} specification types remain the most challenging specification types to transfer to.

\begin{figure}[h!!!]
\centering
     \begin{subfigure}[b]{0.2\textwidth}
         \centering
         \includegraphics[width=\textwidth]{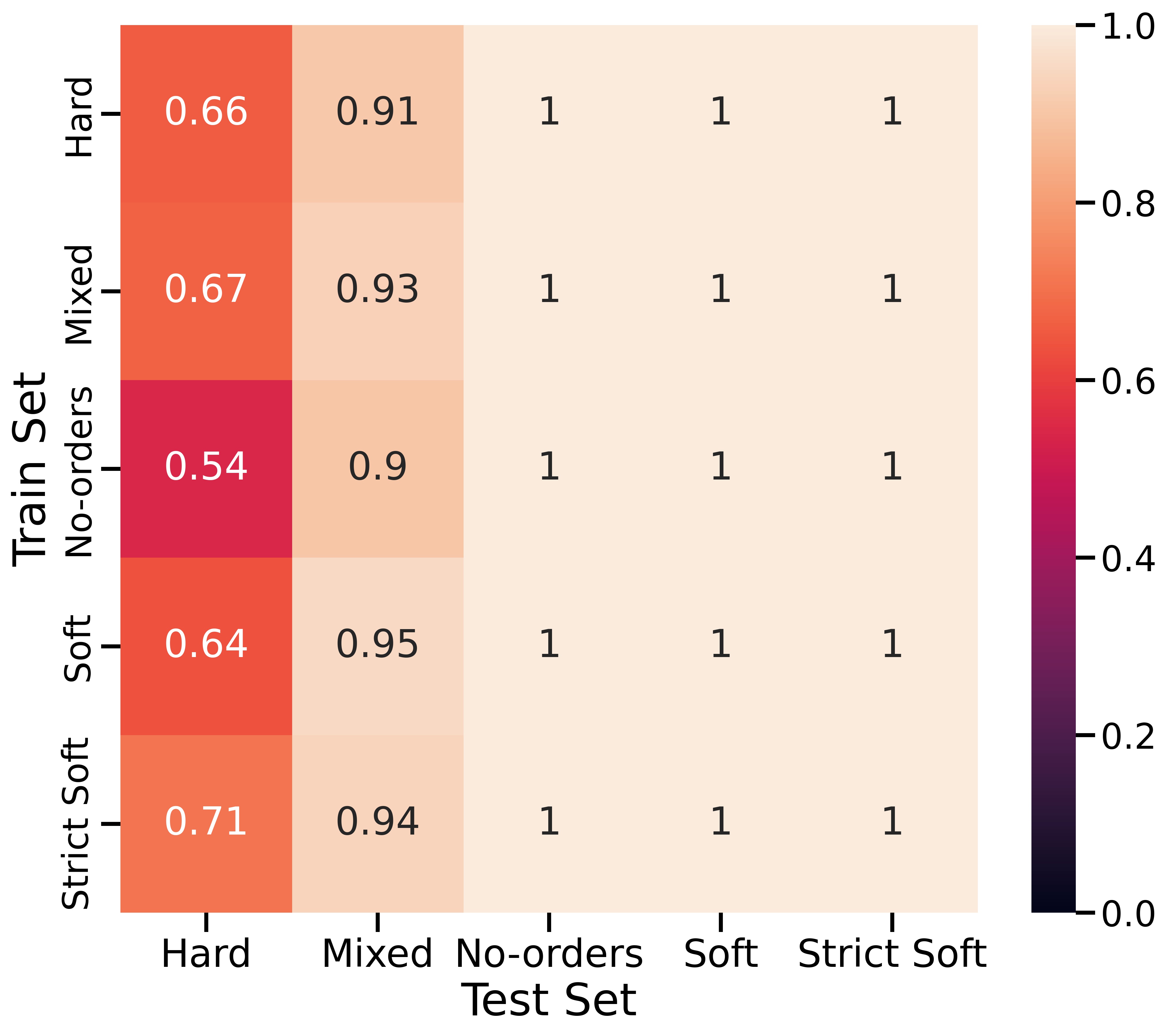}
         \caption{\textit{Relaxed} Edge Match}
         \label{fig:heatmap_relaxed}
     \end{subfigure}
     \begin{subfigure}[b]{0.2\textwidth}
         \centering
         \includegraphics[width=\textwidth]{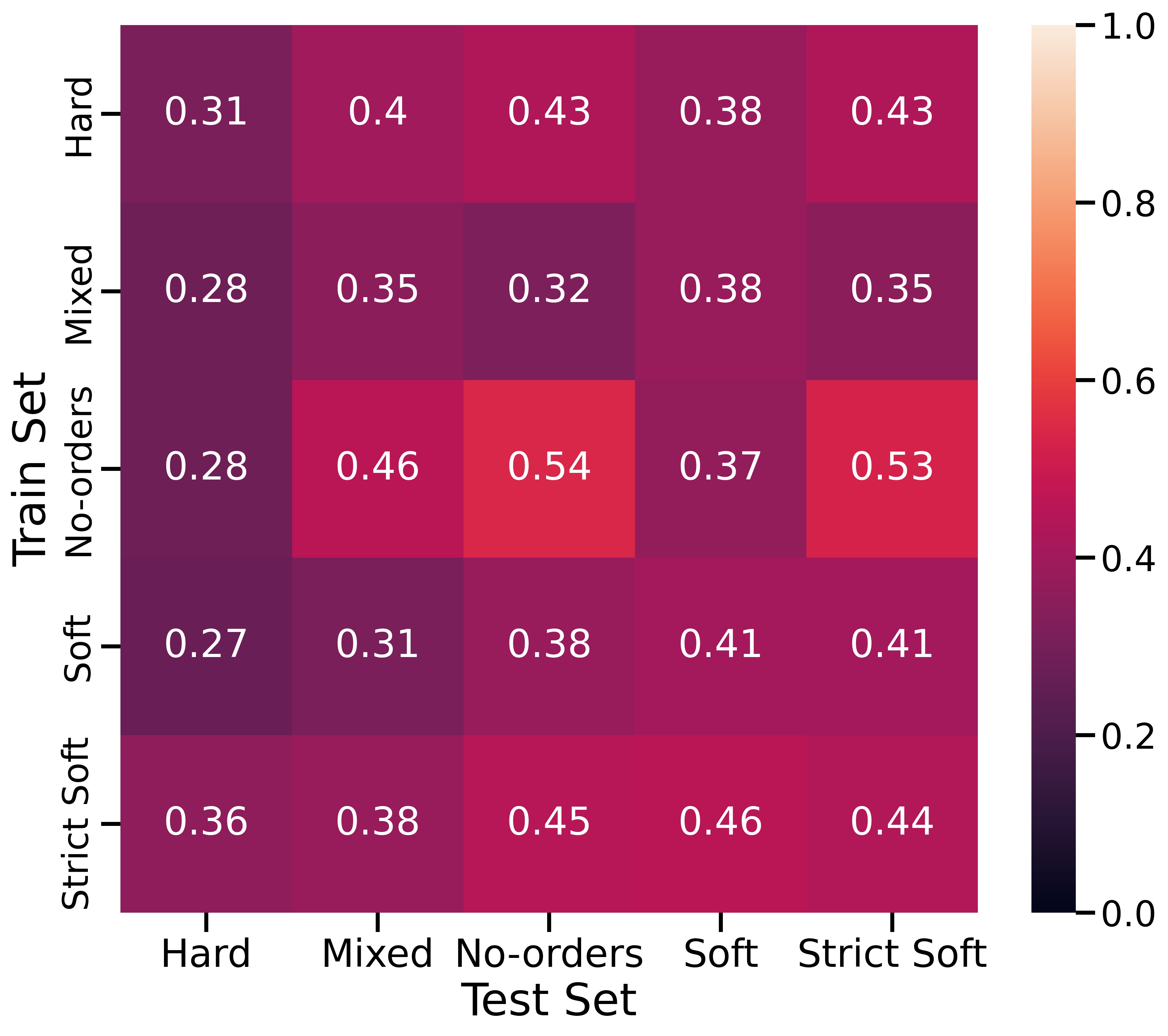}
         \caption{\textit{Constrained} Edge Match}
         \label{fig:heatmap_constrained}
     \end{subfigure}
\caption{
Heatmaps of success rates of LTL-Transfer using the \textit{Relaxed} and \textit{Constrained} edge-matching conditions, respectively, on various training and test specification types.
}
\label{}
\end{figure}

\textbf{Transferability of Specification Types:} We evaluated whether policies learned on different specification types are more capable of transferring to all other specification types.
Figure~\ref{fig:heatmap_relaxed} depicts the heatmap of success rates obtained by training on 50 specifications of the type indicated by the row and transferring to 100 specifications of the type indicated by the column while using the \textit{Relaxed} edge-matching condition.
Similarly, Figure~\ref{fig:heatmap_constrained} depicts the success rates using the \textit{Constrained} edge-matching condition.
No single specification type proved to be the best training set, thus providing evidence against the hypothesis that training with formulas conforming to certain formula templates leads to a greater success rate when transferring to all specification types.
Further, training on all specification types leads to perfect transfer performance on \textit{Soft}, \textit{Strictly-Soft}, and \textit{No-Orders} test set, thus providing further evidence for \textbf{H3} in the main paper.

\textbf{Failure Analysis}: As described in the main paper, we logged the reason for the failure of each unsuccessful transfer attempt.
There are three possible causes: 
\begin{enumerate}
    \item \textit{Specification failure}: A constraint is violated during execution, and the reward machine progresses to an unrecoverable state.

    \item \textit{No feasible path}: After pruning the reward machine graph by removing infeasible edges, there are no paths connecting the start state to an accepting state with matching transition-centric options.

    \item \textit{Options exhausted}: During transferring, there are no further transition-centric options available to further progress the state of the reward machine.
\end{enumerate} 

\begin{figure}[!!!ht]
\centering
    \begin{subfigure}[b]{0.23\textwidth}
         \centering
         \includegraphics[width=\textwidth]{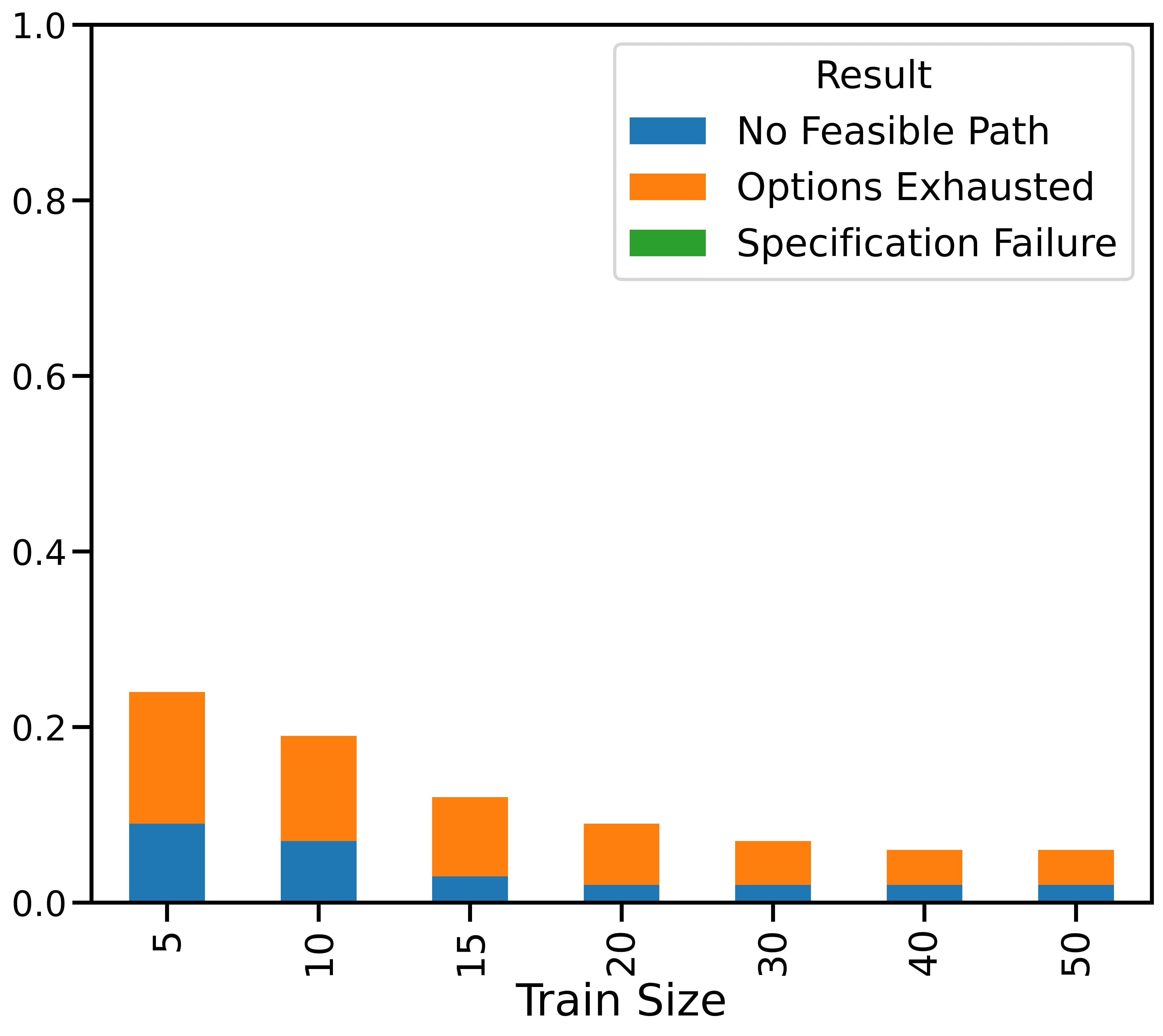}
         \caption{\textit{Relaxed} Edge Match}
         \label{fig:fail_relaxed}
     \end{subfigure}
     \begin{subfigure}[b]{0.23\textwidth}
         \centering
         \includegraphics[width=\textwidth]{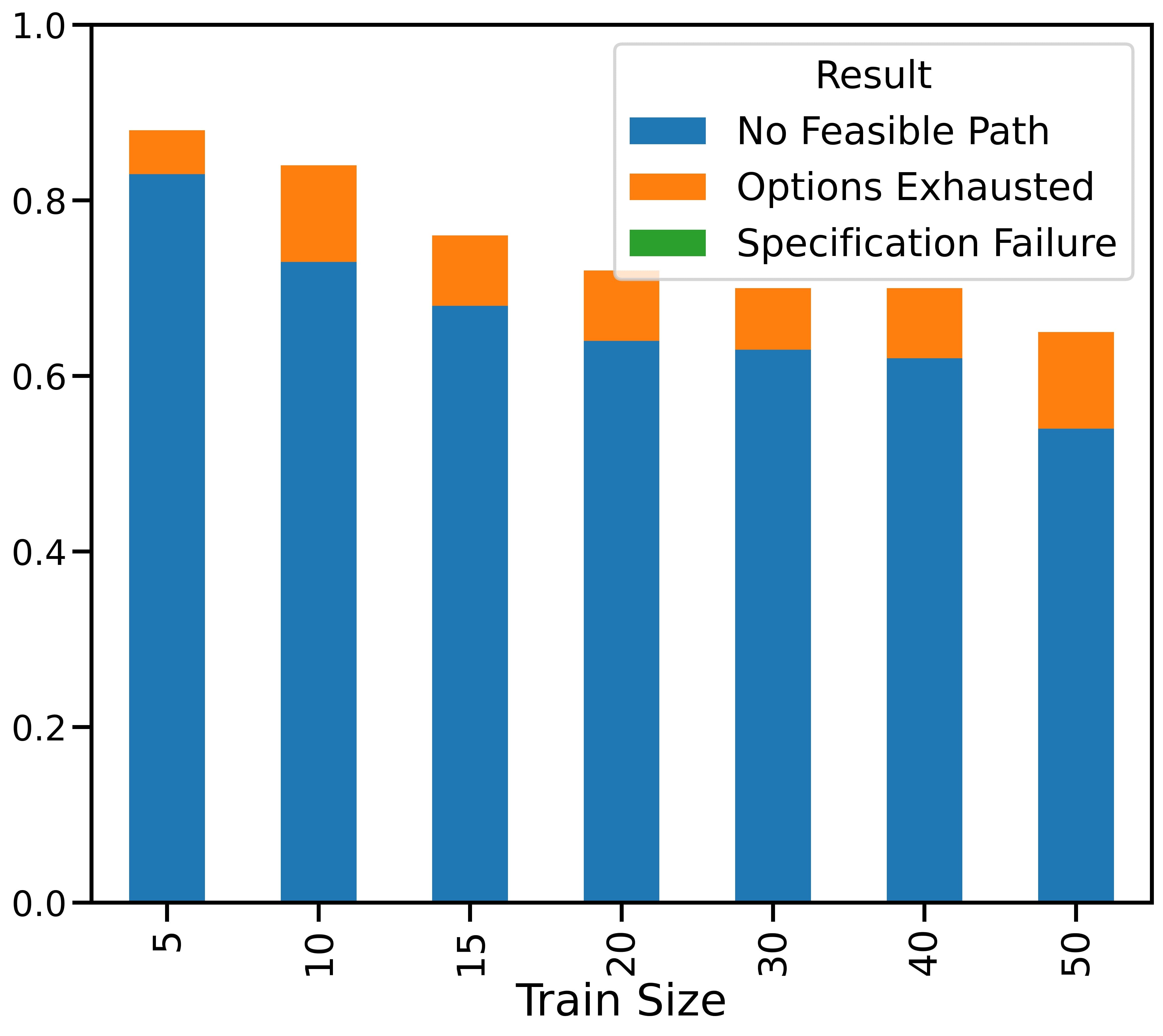}
         \caption{\textit{Constrained} Edge Match}
         \label{fig:fail_constrained}
     \end{subfigure}
     \caption{Reasons for failed task execution after being trained and evaluated on the \textit{Mixed} task specification datasets. Note that all values are depicted in fractions.}
     \label{fig:failure_mode}
\end{figure}

Figure~\ref{fig:failure_mode} depicts the relative frequency of the failure modes when LTL-Transfer is trained and tested on~\textit{mixed} task specifications.
Note that with the~\textit{Relaxed} edge-matching condition not progressing the task after utilizing all available safe options is the primary reason for failure (Figure~\ref{fig:fail_relaxed}) whereas, with the~\textit{Constrained} edge-matching condition, the absence of feasible paths connecting the start and the accepting state is the primary reason for failure (Figure~\ref{fig:fail_constrained}).
Neither has specification failures due to violating a safety constraint.

\section{Selected Solution Trajectories in Simulation}
Consider the case with~\textit{Mixed} training set with $5$ formulas on \textit{Map~0}.
The training formulas are:
\begin{itemize}
    \item $ \F grass ~\wedge ~ \F shelter ~ \wedge ~ \F(wood ~\wedge ~\X \F workbench)$
    \item $\F toolshed ~\wedge ~\F workbench ~ \wedge ~ \F shelter ~\wedge ~ \\(\neg toolshed ~\U ~shelter) ~\wedge~ \F(grass ~\wedge ~\F bridge)$
    \item $\F toolshed ~\wedge ~ \F(shelter ~\wedge ~ \F (axe ~ \wedge ~ \F wood))$
    \item $\F iron ~\wedge~ \F(shelter ~\wedge ~ \X \F(bridge ~\wedge ~ \X \F factory))$
    \item $\F factory$
\end{itemize}

\begin{figure*}[h!]
\centering
     \begin{subfigure}[b]{0.45\textwidth}
         \centering
         \includegraphics[width=\textwidth]{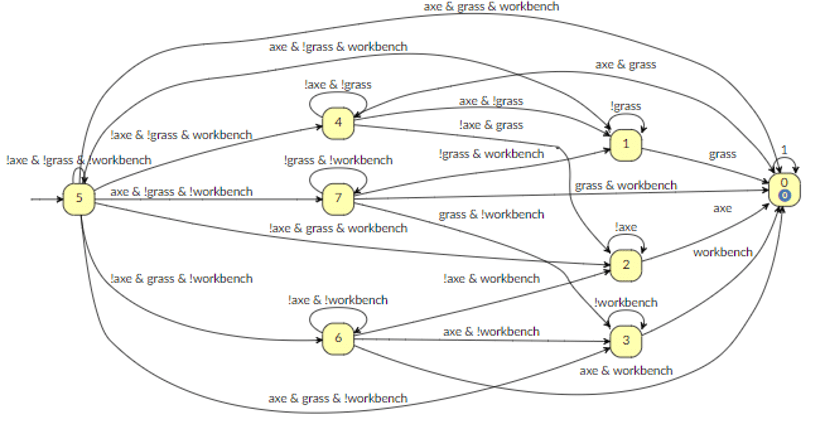}
         \caption{}
         \label{fig:rm_original}
     \end{subfigure}
     \begin{subfigure}[b]{0.45\textwidth}
         \centering
         \includegraphics[width=\textwidth]{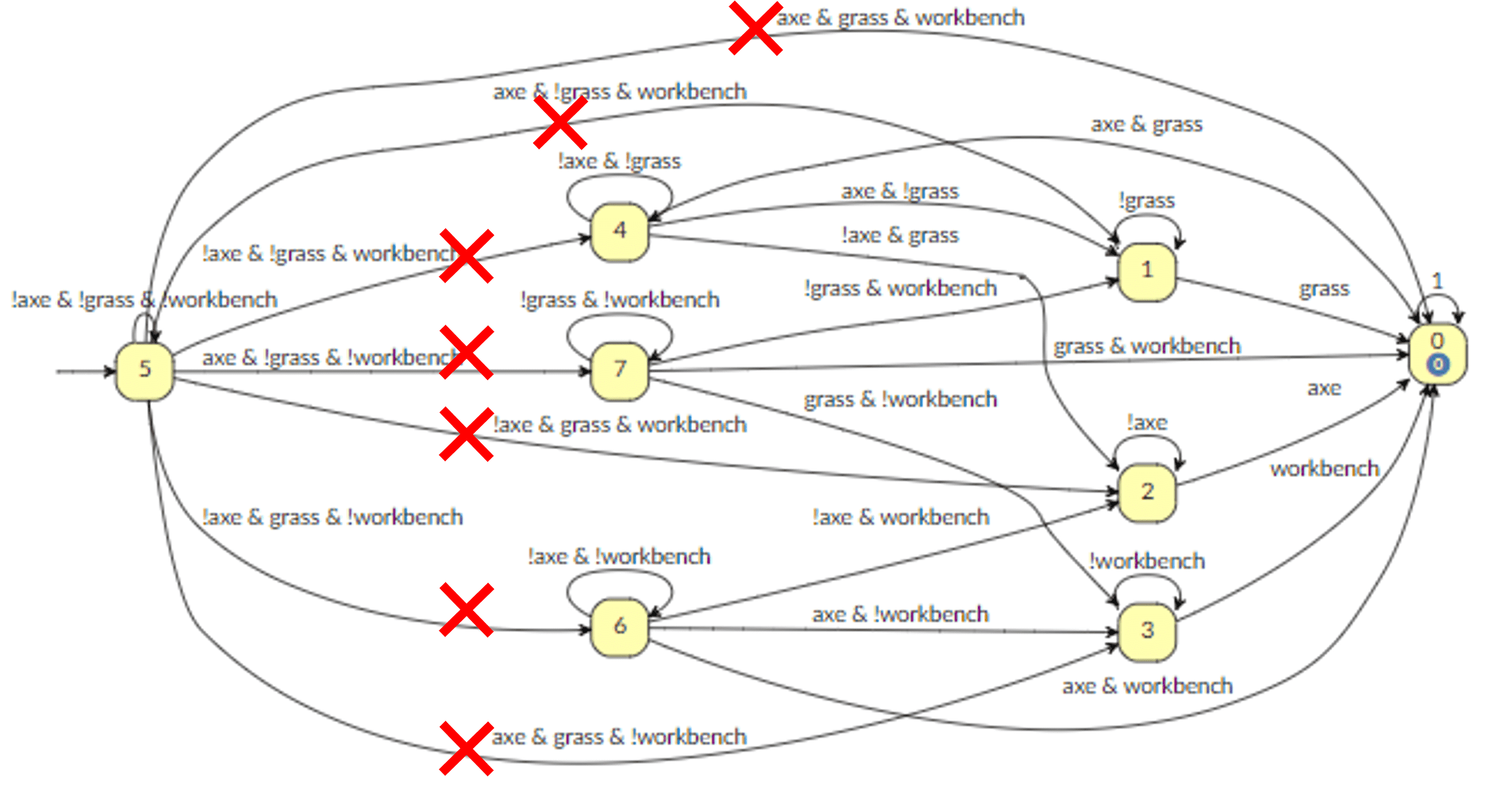}
         \caption{}
         \label{fig:rm_pruned}
     \end{subfigure}
     \caption{
     Figure~\ref{fig:rm_original} depicts the reward machine graph for the task specification $\varphi_{test} = \F workbench ~ \wedge ~ \F grass ~\wedge ~ \F axe$, as well as all feasible edges matched by the \textit{Relaxed} edge-matching condition.
     Note that all the edges have at least one matching transition-centric option.
     Figure~\ref{fig:rm_pruned} depicts the edges that do not have a compatible transition-centric option for the \textit{Constrained} edge-matching condition.
     }
     \label{fig:RM}
\end{figure*}

One of the~\textit{Mixed} test formulas is $\varphi_{test} = \F workbench ~ \wedge ~ \F grass ~\wedge ~ \F axe$. The reward machine for this task specification is depicted in Figure~\ref{fig:rm_original}. 
Given the training set of formulas and the use of the \textit{Constrained} edge-matching condition, the start reward machine state is disconnected from all downstream states as no transition-centric options match the edge transitions, shown in Figure~\ref{fig:rm_pruned}. 
Therefore, LTL-Transfer does not attempt to solve the task and returns failure with the reason being~\textit{no feasible path}, i.e., a disconnected reward machine graph after removing infeasible edges. 

\begin{figure*}[!!!ht]
    \centering
    \includegraphics[width = 0.6\textwidth]{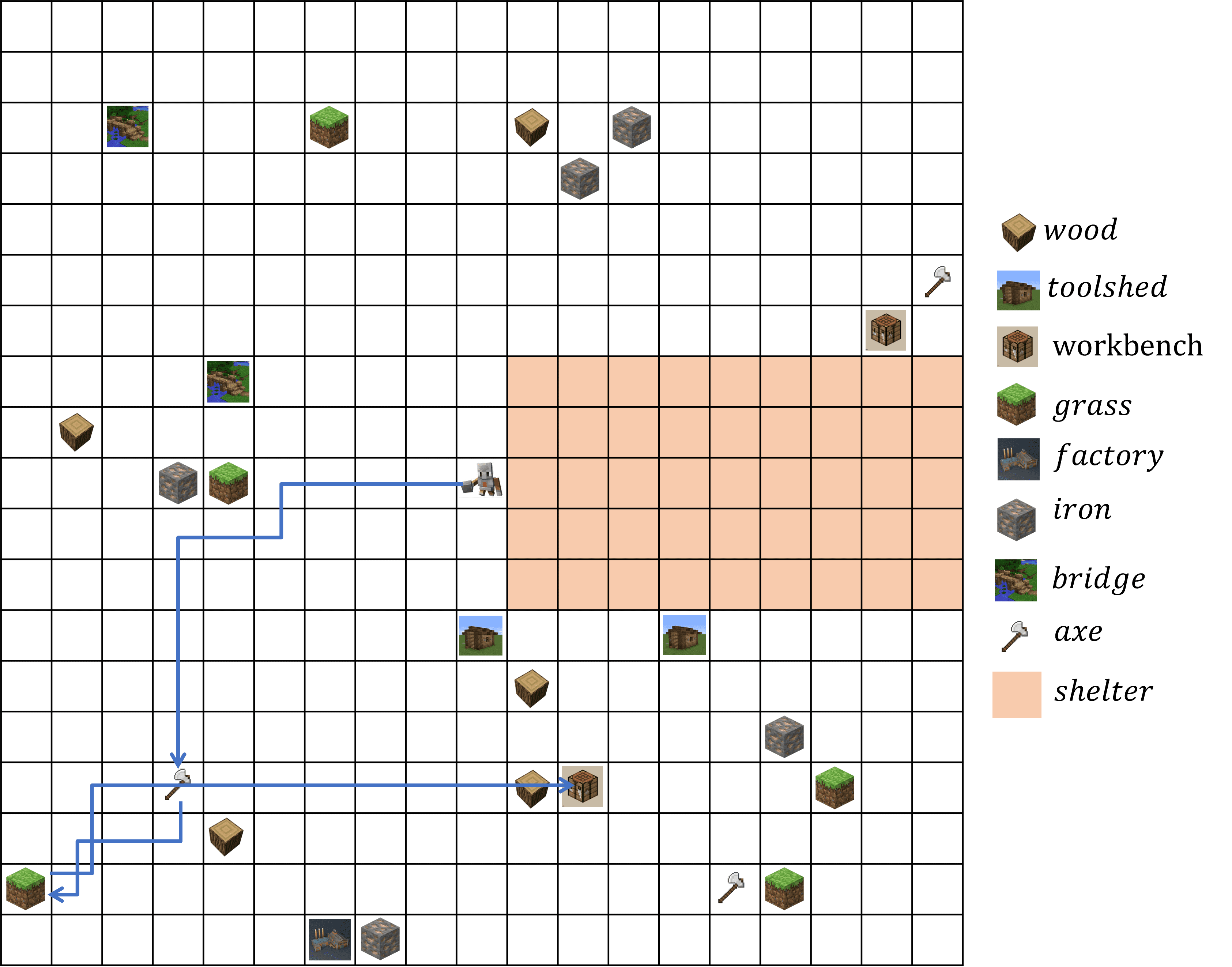}
    \caption{
    Trajectory executed by the robot using LTL-Transfer to achieve the novel task specification $\varphi_{test}= \F workbench ~\wedge ~ \F grass ~\wedge ~ \F axe$.
    }
    \label{fig:traj}
\end{figure*}

If the~\textit{Relaxed} edge-matching condition is used, there are matching transition-centric options for each edge, shown in Figure~\ref{fig:rm_original}.
The trajectory produced by LTL-Transfer when transferring the policies is depicted in Figure~\ref{fig:traj}.
The robot collects all three requisite resources before it terminates the task execution.
Further, note that the robot passes through a grid containing $wood$ as the specification does not explicitly prohibit it.

\section{Robot Demonstration}
The 50 test tasks executed on the robot are shown in Table~\ref{tab:test_tasks}.
The Proposition $a$ represents a brown desk, $b$ represents a white desk, $c$ represents a couch, $d$ represents a door, $s$ represents a bookshelf, and $k$ represents a kitchen counter.

\begin{table*}[ht!]
\scriptsize
\caption{Test Tasks Executed on the Robot}
\label{tab:test_tasks}
    \begin{tabular}{lp{0.2\textwidth}p{0.2\textwidth}}
\toprule
LTL Task Specification & Type of Task & Results \\ \midrule

0. $\F a$ & navigation & success \\ 
1. $\F a \wedge \F b$ & navigation & success \\
2. $\F a \wedge \F b \wedge \F c$ & navigation & success \\
3. $\F a \wedge \F b \wedge \F s$ & navigation & success \\
4. $\F a \wedge \F b \wedge \F k$ & navigation & success \\
5. $\F a \wedge \F b \wedge \F c \wedge \F d$ & navigation & success \\
6. $\F a \wedge \F b \wedge \F c \wedge \F s$ & navigation & success \\
7. $\F a \wedge \F b \wedge \F c \wedge \F k$ & navigation & success \\
8. $\F a \wedge \F b \wedge \F c \wedge \F k \wedge \F s$ & navigation & success \\
9. $\F(b \wedge \F(a \wedge \F(c \wedge \F d))))$ & navigation & success \\
10. $\F(s \wedge \F a)$ & fetch and deliver & success \\
11. $\F(s \wedge \F b)$ & fetch and deliver & success \\
12. $\F(a \wedge \F b)$ & navigation & success \\
13. $\F(b \wedge \F a)$ & navigation & success \\
14. $\F(a \wedge \F(s \wedge \F c))$ & fetch and deliver & success \\
15. $\F(b \wedge \F(s \wedge \F c))$ & fetch and deliver & success \\
16. $\F(s \wedge \F(a \wedge \F c))$ & fetch and deliver & success \\
17. $\F(a \wedge \F(b \wedge \F c))$ & navigation & success \\
18. $\F(s \wedge \F(a \wedge \F(k \wedge \F a))))$ & fetch and deliver & success \\
19. $\F(a \wedge \F(b \wedge \F(c \wedge \F d))))$ & navigation & success \\
20. $\F(s \wedge \X\F a)$ & fetch and deliver & success \\
21. $\F(b \wedge \X\F s)$ & navigation & success \\
22. $\F(a \wedge \X\F b)$ & navigation & success \\
23. $\F(b \wedge \X\F a)$ & navigation & success \\
24. $\F(a \wedge \X\F(b \wedge \X\F c))$ & navigation & success \\
25. $\F(a \wedge \X\F(s \wedge \X\F b))$ & fetch and deliver & success \\
26. $\F(b \wedge \X\F(s \wedge \X\F a))$ & fetch and deliver & success \\
27. $\F(s \wedge \X\F(b \wedge \X\F a))$ & fetch and deliver & success \\
28. $\F(k \wedge \X\F b)$ & fetch and deliver & success \\
29. $\F(k \wedge \X\F a)$ & fetch and deliver & success \\
30. $\neg a \U s \wedge \F a$ & fetch and deliver & success \\
31. $\neg b \U a \wedge \F b$ & navigation & success \\
32. $\neg a \U b \wedge \F a$ & navigation & success \\
33. $b \U a \wedge \neg c \U b \wedge \F c$ & navigation & success \\
34. $b \U k \wedge \F b$ & fetch and deliver & success \\
35. $\neg b \U c \wedge \F b$ & navigation & success \\
36. $\neg a \U s \wedge \neg b \U a \wedge \F b$ & fetch and deliver & success \\
37. $\neg s \U a \wedge \neg b \U s \wedge \F b$ & fetch and deliver & success \\
38. $\neg b \U a \wedge \neg s \U b \wedge \F s$ & navigation & success \\
39. $\neg a \U b \wedge \neg s \U a \wedge \F s$ & navigation & success \\
40. $\F a \wedge \F(b \wedge \F c)$ & navigation & success \\
41. $\F a \wedge \neg c \U b \wedge \F c$ & navigation & success \\
42. $\F(a \wedge \F b) \wedge \neg c \U a \wedge \F c$ & navigation & success \\
43. $\F a \wedge \F(b \wedge \X\F c)$ & navigation & success \\
44. $\F(a \wedge \F b) \wedge \neg c \U b \wedge \F c$ & navigation & success \\
45. $\F(b \wedge \F a) \wedge \neg c \U b \wedge \F c$ & navigation & success \\
46. $\F c \wedge (\neg s \U a \wedge \F s)$ & navigation & success \\
47. $\F c \wedge (\neg a \U s \wedge \F a)$ & navigation & success \\
48. $\F c \wedge (\neg s \U b \wedge \F s)$ & navigation & success \\
49. $\F c \wedge (\neg b \U s \wedge \F b)$ & navigation & success \\

\\ \bottomrule
\end{tabular}
\end{table*}

\end{appendices}

\end{document}